%% file: main.tex
\documentclass[acmsmall,nonacm,screen,11pt]{acmart}

\setcopyright{none}
\acmYear{2026}

\usepackage{brand}
\usepackage{header}
\usepackage{docspec}

\usepackage{amsfonts}
\usepackage{amssymb}
\usepackage{pifont}
\AtBeginDocument{\renewcommand{\checkmark}{\text{\ding{51}}}}
\newcommand{\cmark}{\ding{51}}
\newcommand{\xmark}{\ding{55}}
\usepackage{nicefrac}
\usepackage{subcaption}
\usepackage{makecell}
\usepackage{colortbl}
\IfFileExists{algorithm.sty}{%
  \usepackage{algorithm}%
  \usepackage{algorithmic}%
}{%
  \newfloat{algorithm}{htbp}{loa}
  \floatname{algorithm}{Algorithm}
  \newenvironment{algorithmic}[1][0]{\begin{enumerate}[leftmargin=2.4em, itemsep=1pt, label={\footnotesize\arabic*:}]}{\end{enumerate}}
  \newcommand{\STATE}{\item}
  \newcommand{\IF}[1]{\item \textbf{if} ##1 \textbf{then}}
  \newcommand{\ELSE}{\item \textbf{else}}
  \newcommand{\ENDIF}{\item \textbf{end if}}
  \newcommand{\FOR}[1]{\item \textbf{for} ##1 \textbf{do}}
  \newcommand{\ENDFOR}{\item \textbf{end for}}
}
\usepackage{tabularx}
\usepackage{multirow}
\usepackage{enumitem}
\usepackage{tikz}
\usetikzlibrary{positioning, calc, backgrounds, fit, arrows.meta}
\usepackage{pgfplots}
\pgfplotsset{compat=1.18}

\newif\ifcomments
\commentstrue
\ifcomments
  \usepackage[textsize=footnotesize, textwidth=1.7cm]{todonotes}
\else
  \usepackage[disable]{todonotes}
\fi
\makeatletter
\@ifundefined{comment}{}{}
\makeatother

\newcommand{\methodname}{TerraZero}

\begin{document}

\title[\methodname: Procedural Driving Simulation for Zero-Demonstration Self-Play at Scale]{\methodname: Procedural Driving Simulation\\for Zero-Demonstration Self-Play at Scale}

\author{Zhouchonghao Wu}
\affiliation{\institution{Applied Intuition}\city{Mountain View}\state{CA}\country{USA}}
\author{Akshay Rangesh}
\affiliation{\institution{Applied Intuition}\city{Mountain View}\state{CA}\country{USA}}
\author{Weixin Li}
\affiliation{\institution{Applied Intuition}\city{Mountain View}\state{CA}\country{USA}}
\author{Wei-Jer Chang}
\affiliation{\institution{Applied Intuition}\city{Mountain View}\state{CA}\country{USA}}
\author{Zachary Lee}
\affiliation{\institution{Applied Intuition}\city{Mountain View}\state{CA}\country{USA}}
\author{Saeed Bonab}
\affiliation{\institution{Applied Intuition}\city{Mountain View}\state{CA}\country{USA}}
\author{Tim Wang}
\affiliation{\institution{Applied Intuition}\city{Mountain View}\state{CA}\country{USA}}
\author{Wei Zhan}
\affiliation{\institution{Applied Intuition}\city{Mountain View}\state{CA}\country{USA}}

\compactauthors{%
  {\raggedright
   {\rmfamily\bfseries
    Zhouchonghao Wu\textsuperscript{1,*}, \;
    Akshay Rangesh\textsuperscript{1,*}, \;
    Weixin Li\textsuperscript{1}, \;
    Wei-Jer Chang\textsuperscript{1,2},\\[0.25em]
    Zachary Lee\textsuperscript{1}, \;
    Saeed Bonab\textsuperscript{1}, \;
    Tim Wang\textsuperscript{1}, \;
    Wei Zhan\textsuperscript{1,2,\dag}\par}
   \smallskip
   {\small\rmfamily\color{AppliedGray}%
    \textsuperscript{1}Applied Intuition \quad
    \textsuperscript{2}University of California, Berkeley\par}
   \smallskip
   {\footnotesize\rmfamily\color{AppliedGray}%
    \textsuperscript{*}Equal contribution. \quad
    \textsuperscript{\dag}Corresponding author: wei.zhan@applied.co\par}
   \smallskip
   {\footnotesize\rmfamily\color{AppliedGray}%
    Project website: \url{https://terra-applied.github.io/TerraZero}\par}
   \par}
}

\input{sections/abstract}

\maketitle

\clearpage
\begingroup
\hypersetup{linkcolor=AppliedBlack}
\tableofcontents
\endgroup
\clearpage

\input{sections/introduction}

\begin{figure}[t]
\centering
\includegraphics[width=\linewidth]{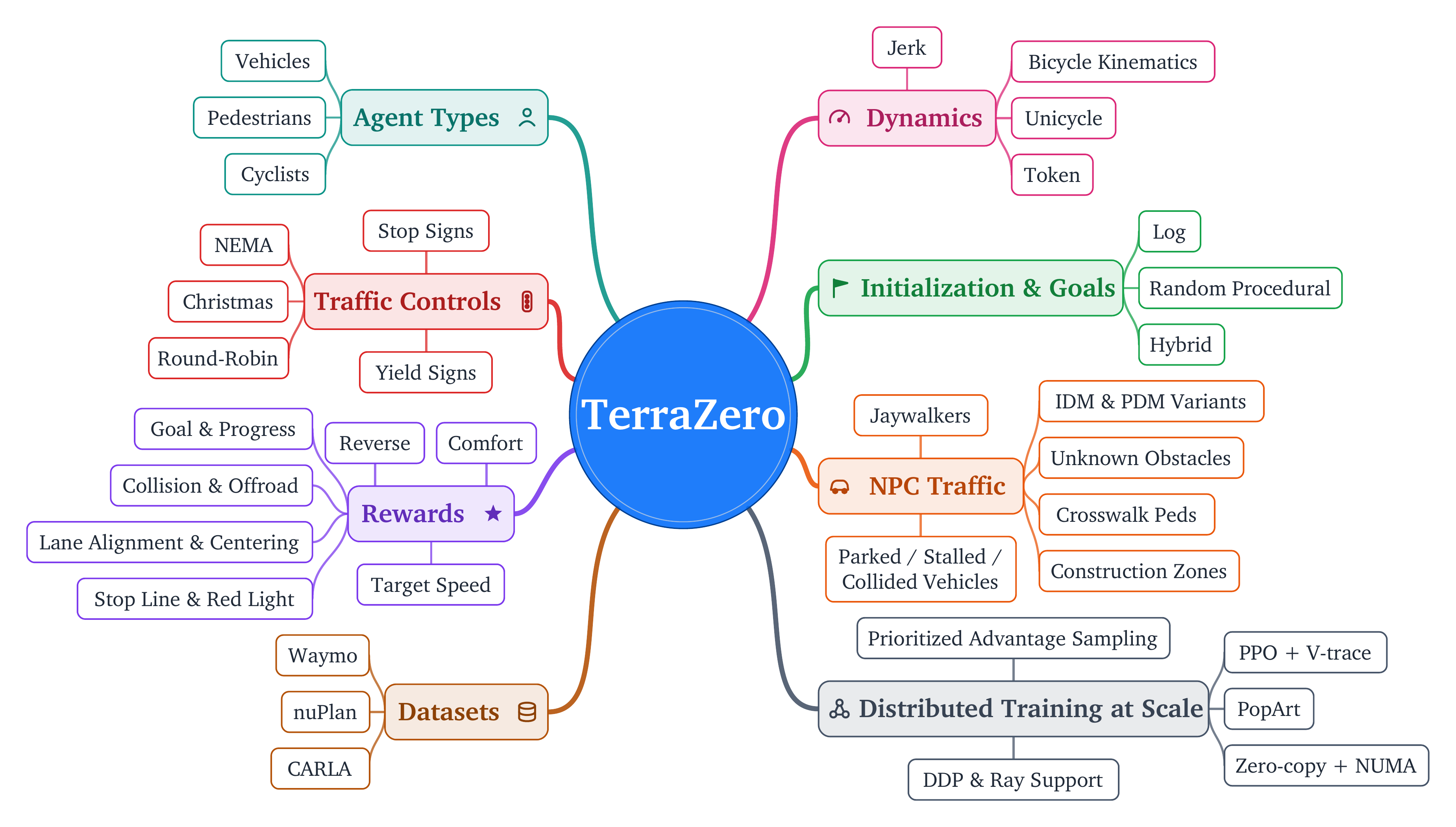}
\caption{\textbf{\methodname{} at a glance.} \methodname{} couples three pillars: a fast object-level simulator, a procedural scenario generator, and a self-play training recipe. Each branch is an axis the system can control: how agents move, who shares the road, how episodes begin, the rule-based traffic and signal controllers that populate the scene, the reward terms, the source datasets, and the machinery that scales training across GPUs. A configuration fixes some of these choices and randomizes the rest across episodes, which is how \methodname{} manufactures hard scenarios from real-world map data.}
\label{fig:mindmap}
\end{figure}
\input{sections/related_work}
\input{sections/system}
\input{sections/scenario_generation}

\input{sections/training}
\input{sections/experiments}
\input{sections/conclusion}

\bibliographystyle{ACM-Reference-Format}
\bibliography{references}

\newpage
\appendix
\input{sections/appendix}

\end{document}

%% file: sections/abstract.tex
\begin{abstract}
Training robust autonomous driving agents requires a simulator that is fast enough for reinforcement learning at scale, realistic enough to ground behavior in real-world map structure, and diverse enough to cover the safety-critical long tail that logged driving data rarely contains.
We present \textbf{\methodname}, a procedural driving simulator and self-play training stack that meets these goals.
A configurable C engine runs simulation on the CPU and policy inference on the GPU over a zero-copy data path, sustaining 1.3M agent-steps per second on a single server-grade GPU, significantly faster than existing object-level simulators, while retaining fidelity that lighter single-agent systems omit: heterogeneous agents, multiple dynamics models, and full traffic-rule enforcement.
Rather than treating logged data as the training distribution, \methodname{} uses it only as a source of real-world map geometry: it populates each map with a randomized cast of rule-based road users and signal controllers, and randomizes agent dynamics, rewards, and sizes per episode, so a single map yields an effectively unbounded supply of training scenarios.
\methodname{} trains every reported policy from scratch by reinforcement learning alone, with \emph{zero} human demonstrations. It uses no imitation, no logged trajectories, and no fallback planner at inference, on a compute-efficient self-play recipe scaled across GPUs. The resulting policies generalize \emph{zero-shot} across cities and datasets, including emergent left-hand-traffic driving without explicit supervision.
As an ego policy, a single checkpoint is, to our knowledge, the first fully learned policy to top both the standard val14 split and the interactive long-tail InterPlan suite.
On Waymo Open Sim Agents realism the same recipe yields a sim agent that outperforms other demonstration-free methods and is competitive with the strongest reference-anchored self-play method.
A single stack therefore serves both roles: it trains state-of-the-art demonstration-free driving policies across vehicle dynamics suited to different platforms such as cars and trucks, and sim agents that jointly control the full traffic set of vehicles, pedestrians, and cyclists, all with configurability broad enough to meet a wide range of needs.
\end{abstract}

%% file: sections/introduction.tex
\section{Introduction}
Real-world driving data is dominated by routine driving.
Most logged miles consist of steady-state lane following, gentle curves, and routine stops, the scenarios an autonomous agent has least to gain from practicing.
The long tail of safety-critical situations that largely determine deployment readiness (dense highway merges, aggressive cut-ins, near-miss pedestrian interactions, unprotected left turns through oncoming traffic) is rare, and therefore costly to cover, in real-world driving datasets.
Imitation learning on logged trajectories struggles to teach behaviors that the logs barely contain, and collecting more data at fleet scale closes this distributional gap only slowly.
Simulation offers a complementary path: a simulator can systematically manufacture the hard scenarios that matter most~\citep{dosovitskiy2017carla, vinitsky2022nocturne, kazemkhani2025gpudrive}.

Existing simulation approaches occupy different points in a throughput--fidelity trade-off, and none fully resolves it.
Object-level simulators built for speed, such as PufferDrive~\citep{pufferdrive2025github,cornelisse2025reliable}, reach high throughput but support only a single agent type with fixed dynamics and no traffic signals.
Feature-rich environments like SMARTS~\citep{zhou2021smarts} and CARLA~\citep{dosovitskiy2017carla} offer the complexity needed for realistic evaluation but are orders of magnitude too slow for from-scratch reinforcement learning (RL) training at scale, where billions of environment steps are routine.
A separate line of work builds learned traffic models (SimNet~\citep{bergamini2021simnet}, BITS~\citep{xu2023bits}, Trajeglish~\citep{philion2024trajeglish}) that generate realistic trajectories for surrounding agents but produce largely non-reactive traffic, limiting their utility for closed-loop policy improvement.
Self-play approaches like Gigaflow~\citep{cusumano-towner2025gigaflow} show that reactive multi-agent training can yield robust driving policies, but they train on a small family of synthetic maps under a single dynamics model shared by every agent class, and they rely on randomization alone, rather than constructed scenario content, for exposure to the long tail.

\methodname{} closes this gap through \emph{procedural scenario generation}.
It treats logged data as a source of real-world map geometry, and composes a stack of randomization mechanisms to manufacture hard scenes from each map, while a configurable C simulation engine supplies the throughput that large-scale RL demands.
Each map is populated with a procedurally generated cast of rule-based road users (parked and reactive vehicles, planner-driven traffic, crashed-vehicle clusters, construction zones, and crossing and jaywalking pedestrians), driven through randomized traffic-signal controllers, with per-episode randomization of agent dynamics~\citep{tobin2017domainrandomization}, reward weights, and bounding-box sizes layered on top.
Because every axis varies from episode to episode, one map seeds a vast and non-repeating space of training scenarios.
The maps themselves span three datasets: the Waymo Open Motion Dataset (WOMD)~\citep{ettinger2021womd}, nuPlan~\citep{caesar2021nuplan} across four cities that span both right- and left-hand traffic, and five synthetic CARLA~\citep{dosovitskiy2017carla} towns, giving broad coverage of intersection layouts, road topologies, and driving conventions.
\methodname{} provides a complete training stack from scenario loading through distributed Proximal Policy Optimization (PPO) and evaluation.
Figure~\ref{fig:mindmap} summarizes the configuration surface and the training stack.

Our contributions are threefold:
\begin{enumerate}[leftmargin=*, itemsep=2pt, topsep=4pt]
    \item A \textbf{fast, feature-rich object-level traffic simulator} for RL training. A configurable C engine runs simulation on the CPU and policy inference on the GPU, connected by a zero-copy data path with dense feature packing, 16-bit observations, non-uniform memory access (NUMA)-aware orchestration, and a compact binary scenario format. It supports heterogeneous agents (vehicles, pedestrians, cyclists), multiple dynamics models, and full traffic-rule enforcement, sustaining up to 2.8M agent-steps per second (SPS) on an 8-GPU node, to our knowledge significantly faster than any existing driving simulator, while retaining a fidelity that lighter single-agent systems omit.
    \item A \textbf{procedural scenario generator} that treats logged data as the starting distribution rather than the training distribution. From real-world maps it manufactures diverse scenarios through randomized initialization and goal sampling, agent density and dimension randomization, a configurable population of rule-based road users, and three traffic-signal controllers, so a single map yields an effectively unbounded supply of training scenarios.
    \item A \textbf{robust self-play training recipe} for object-level driving policies. Because the simulator is fast, the recipe trades sample efficiency for compute efficiency via saliency-prioritized sampling, and combines a compact feed-forward policy with V-trace off-policy corrections, PopArt value normalization, reward and kinematic domain randomization, and population play that breaks self-play symmetry, scaled across GPUs with synchronized normalization statistics. The \emph{zero} in \methodname{} names the training stance: zero human demonstrations. Because the recipe learns from scratch by reinforcement signal alone, it needs no imitation loss and no logged trajectories, and every result we report is trained this way. Log and hybrid initialization from logged data stay available as options that our reported policies do not use. The resulting policies generalize zero-shot across cities and datasets, including emergent left-hand-traffic driving without explicit supervision.
\end{enumerate}
We validate these contributions on three public datasets (Waymo, nuPlan, CARLA), measuring driving performance on nuPlan val14 and InterPlan and sim-agent realism on the Waymo Open Sim Agents Challenge (WOSAC).
Across the three, one recipe tops both the standard val14 benchmark and the interactive long-tail InterPlan benchmark, and outperforms other demonstration-free methods on WOSAC realism, evidence that one stack can serve both traffic simulation and high-performance planning.

%% file: sections/related_work.tex
\section{Related Work}
\label{sec:related_work}

\subsection{Object-Level Driving Simulators}

The choice of simulation abstraction sets the ceiling on reinforcement learning scale. Pixel-level simulators such as CARLA~\citep{dosovitskiy2017carla} render photorealistic urban scenes with full sensor suites, but their rendering cost caps throughput at tens of frames per second. The long-tail situations that most determine deployment readiness are hard to learn, and learning them from scratch through self-play is harder still; both demand orders of magnitude more environment steps than pixel-level rendering can supply. \emph{Object-level} simulators represent the world as structured state (positions, velocities, road geometry) rather than pixels, trading perceptual realism for the throughput that large-scale RL requires. \methodname{} takes this trade.

Nocturne~\citep{vinitsky2022nocturne} introduced object-level driving simulation with a C++ backend over Waymo Open Motion Dataset scenarios, achieving large speedups over pixel-level alternatives. It remains narrow, however: it drives vehicles only, models no traffic signals, and offers a single dynamics model. SMARTS~\citep{zhou2021smarts} is more feature-rich, with multiple agent types and flexible scenario specification, but its Python simulation loop bottlenecks on the CPU and caps throughput for large-scale RL. Hardware acceleration lifts that ceiling. Waymax~\citep{gulino2023waymax} runs batched rollouts in JAX for planning, behavior prediction, and sim-agent research, and GPUDrive~\citep{kazemkhani2025gpudrive} lowers observation, reward, and dynamics functions to CUDA on the Madrona game engine, reporting roughly one million steps per second under heavily batched rollouts. Both stay narrow in scenario content: Waymax centers on the WOMD format and a benchmark-oriented API rather than a distributed self-play stack, while GPUDrive drives vehicles only with no signal modeling, so under a realistic configuration (Table~\ref{tab:simulator_comparison}) its sustained agent throughput falls well below that headline figure. PufferDrive~\citep{pufferdrive2025github,cornelisse2025reliable} is the closest antecedent, a fast C backend over Waymo scenarios, but it too supports a single agent type with limited dynamics.

Gigaflow~\citep{cusumano-towner2025gigaflow} pairs a well-engineered platform with self-play to produce robust driving policies. As a training substrate it carries three limitations. First, it trains on a small family of synthetic CARLA-derived maps, affine variants of one fixed network rather than real-world geometry. Second, its scenes hold only policy-driven agents and static obstacles modeled as untyped immobile vehicles, leaving out the real-world situations that any road-ready driving policy must expect to encounter, such as construction zones, crashed-vehicle clusters, curb-parked cars, and jaywalking pedestrians, the long-tail cases that benchmarks such as InterPlan~\citep{hallgarten2024interplan} probe. Third, it drives every agent class through a single bicycle model, covering trucks as size variants of the same box, and at evaluation hands pedestrians to a scripted controller in the style of the Intelligent Driver Model (IDM)~\citep{treiber2000idm}, so its unified control of pedestrians runs only inside its own simulator. Table~\ref{tab:simulator_comparison} summarizes these distinctions.
\begin{table}[t]
\centering
\caption{Comparison of object-level driving simulators. \methodname{} combines C-level simulation speed with scenario fidelity (heterogeneous agents, multiple dynamics models, traffic signals) typically found only in slower, more feature-rich systems. Agent SPS is reported separately for a single consumer-grade GPU, a single server-grade GPU, and an 8$\times$ server-grade GPU node. Numbers marked $^\dagger$ are figures reported by the original authors; all other SPS values are benchmarked under our setup. ``Multi-node'' indicates whether the system supports distributed training across multiple machines.}
\label{tab:simulator_comparison}
\small
\resizebox{\textwidth}{!}{%
\begin{tabular}{ll ccc c llccc}
\toprule
& & \multicolumn{3}{c}{\textbf{Agent SPS}} & & & & & & \\
\cmidrule(lr){3-5}
\textbf{System} & \textbf{Backend} & \textbf{Consumer} & \textbf{Server} & \textbf{8$\times$ Server} & \textbf{Multi-node} & \textbf{Agent Types} & \textbf{Dyn. Models} & \textbf{Signals} & \textbf{Reactive} & \textbf{Data Sources} \\
\midrule
Nocturne    & C++       & 2K$^\dagger$ & ---  & ---  & ---        & Vehicles      & 1        & ---        & ---        & Waymo \\
SMARTS      & Python    & ---  & ---  & ---  & \checkmark & Veh, Ped, Cyc & Multiple & \checkmark & Partial    & Synthetic (SUMO) \\
Waymax      & JAX       & ---  & 319K$^\dagger$ & ---  & \checkmark & Veh, Ped, Cyc & 2        & \checkmark & \checkmark & Waymo \\
GPUDrive    & Madrona   & 11K  & 22K  & ---  & ---        & Vehicles      & 1        & ---        & ---        & Waymo \\
PufferDrive & C         & 320K & 745K & ---  & ---        & Vehicles      & 1        & ---        & \checkmark        & Waymo \\
Gigaflow    & ---       & ---  & ---  & 530K$^\dagger$ & --- & Veh, Ped, Cyc & 1        & \checkmark & \checkmark & Synthetic (CARLA) \\
\textbf{\methodname} & \textbf{C} & \textbf{560K} & \textbf{1.3M} & \textbf{2.8M} & \checkmark & \textbf{Veh, Ped, Cyc} & \textbf{Multiple} & \checkmark & \checkmark & \textbf{Waymo, nuPlan, CARLA} \\
\bottomrule
\end{tabular}%
}
\end{table}
\subsection{Traffic Simulation via Imitation Learning}

A complementary line of work learns traffic agent behavior directly from logged driving data, producing simulated actors that reproduce the statistical properties of real traffic. SimNet~\citep{bergamini2021simnet} trains a neural network to generate reactive multi-agent rollouts from real-world observations. BITS~\citep{xu2023bits} introduces a bi-level formulation that separates high-level route intentions from low-level trajectory generation. Trajeglish~\citep{philion2024trajeglish} reframes traffic modeling as next-token prediction over discretized trajectory tokens, leveraging the scalability of autoregressive sequence models. CTG++~\citep{zhong2023ctgpp} conditions traffic generation on natural language descriptions via diffusion models. TrafficBots~\citep{zhang2023trafficbots} learns action-conditioned world models that can simulate multi-agent interactions. ITRA~\citep{scibior2021itra} formulates differentiable simulation that enables gradient-based optimization of traffic scenarios.

These methods excel at producing statistically faithful reproductions of logged traffic patterns and have shown strong results on realism benchmarks such as WOSAC~\citep{montali2023wosac}. Related planner and policy benchmarks such as NAVSIM~\citep{dauner2024navsim} instead measure ego-driving performance under non-reactive pseudo-simulation. Their fidelity is bounded by the quality of the logs they learn from: WOMD, for instance, carries sparse traffic-signal annotations and very few cyclists, and a shift in sensor configuration or offboard-labeling accuracy can move the target they reproduce. These models are also optimized for fidelity to the \emph{logged} distribution rather than for generating challenging, interactive scenarios, and most are either non-reactive to the ego agent's decisions or offer only limited reactivity that degrades over long rollout horizons, which leaves them less suitable as training environments for RL.

\subsection{Self-Play for Driving}
\label{subsec:selfplay}

Reinforcement learning for autonomous driving has shown increasing promise as simulation throughput has improved. Early work demonstrated RL-trained agents in simplified settings~\citep{kendall2019learning}, with later efforts carrying the paradigm into domains as varied as off-road terrain~\citep{wu2026tadpo}, and more recent efforts scaling to complex multi-agent scenarios on object-level simulators~\citep{vinitsky2022nocturne, kazemkhani2025gpudrive}. A key challenge is generating sufficiently diverse and challenging training scenarios: logged data provides realistic but mundane distributions, while hand-crafted scenarios are labor-intensive and inevitably incomplete.

Self-play offers a solution. By training an agent against copies of itself or its prior checkpoints, self-play creates an automatic curriculum: as the policy improves, the traffic it trains against becomes proportionally more competent, generating progressively harder interactions without manual design. This principle has driven breakthroughs across domains, from Go~\citep{silver2017alphago} to Dota~\citep{berner2019dota} to multi-agent locomotion~\citep{bansal2018emergent}.

Gigaflow~\citep{cusumano-towner2025gigaflow} gives the most comprehensive demonstration for driving, combining multi-agent self-play with domain randomization to produce policies robust to distributional shift, though it does so entirely on synthetic maps with homogeneous agent dynamics, as discussed above. SPACeR~\citep{chang2026spacer} stabilizes self-play with centralized reference models that anchor agent behavior against the policy drift of unconstrained co-adaptation, at the cost of maintaining a separate reference model. HR-PPO~\citep{cornelisse2024hrppo} regularizes self-play with imitation losses to keep policies human-like, which tethers behavior to the logged demonstration distribution rather than letting self-play explore freely beyond it. BehaviorBenchmark~\citep{behaviorbenchmark2026} contributes a standardized suite for the closed-loop behavior of learned driving policies; we cite its reported InterPlan RL baselines as comparison points rather than running its suite ourselves.

%% file: sections/system.tex
\section{Fast, Feature-Rich Traffic Simulation}
\label{sec:system}

\begin{figure}[t]
\centering
\includegraphics[width=0.8\textwidth]{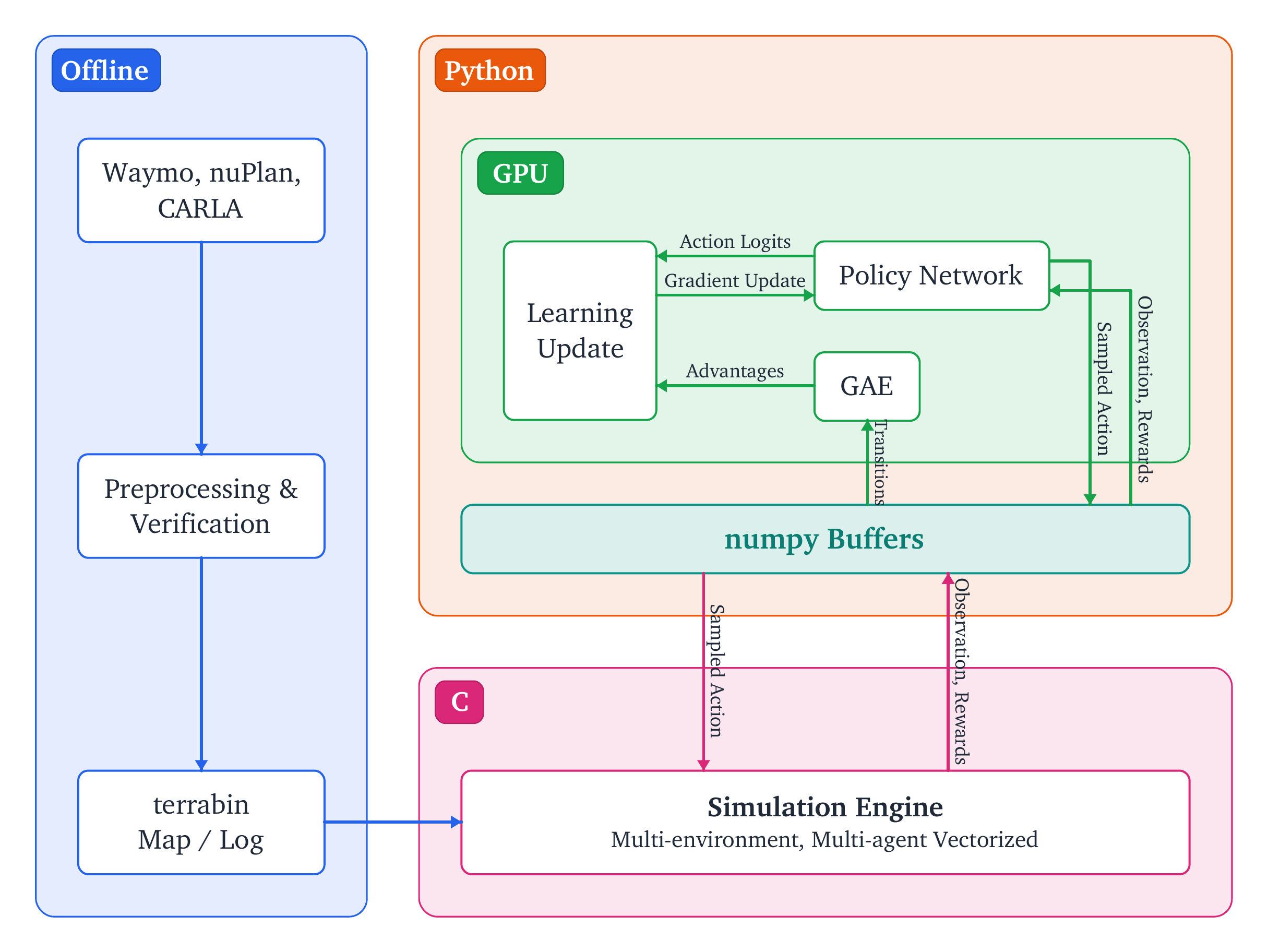}
\caption{\textbf{System architecture.} The offline data pipeline (left) converts Waymo, nuPlan, and CARLA scenarios into a compact binary scenario format (Section~\ref{sec:data_pipeline}). At runtime, the C simulation engine (bottom) exchanges actions and observations with the GPU training loop (top) through shared, zero-copy numpy buffers, with the PPO learning update driving policy improvement.}
\label{fig:system_architecture}
\end{figure}

\methodname{} is built around two goals that are usually in tension: a simulator fast enough for reinforcement learning at scale, and one feature-rich enough to represent the long-tail situations a policy must learn to handle. We achieve both by carefully partitioning work between the CPU and the GPU (Figure~\ref{fig:system_architecture}) and by implementing only the features that measurably improve training outcomes. Appendix~\ref{app:constants} provides a reference table of simulation constants.

\subsection{Designing for Performance}
\label{subsec:performance}

\paragraph{CPU/GPU division of labor.}
The object-level dynamics, observation construction, reward computation, and traffic-light logic run entirely in C, compiled into a CPython extension, while policy inference and learning run in PyTorch on the GPU~\citep{ansel2024pytorch}; the whole stack is built on the PufferLib vectorization and training framework~\citep{suarez2024pufferlib}. Environments execute in parallel worker processes over a shared set of memory buffers, and the trainer consumes the full batch in a single device transfer and batched forward pass. This split keeps the CPU saturated with cheap, branchy simulation work and the GPU saturated with dense tensor math.

\paragraph{Low-overhead data path.}
The interface between the two is fully zero-copy: the C engine writes simulation state directly into the buffers that the training loop reads back as PyTorch tensors, eliminating the per-step serialization and allocation that dominate CPU-bound simulators, while host-to-device transfers overlap with kernel launches. \methodname{} further sizes its buffers to the controlled agents actually present rather than padding to a fixed count, emits observations in 16-bit precision that the GPU reinterprets bit-for-bit to halve observation bandwidth, and pins each rank's workers to the CPU cores local to its GPU's NUMA node.

\subsection{Composable Simulation Substrate}
\label{subsec:substrate}

\begin{figure}[t]
\centering
\includegraphics[width=\textwidth]{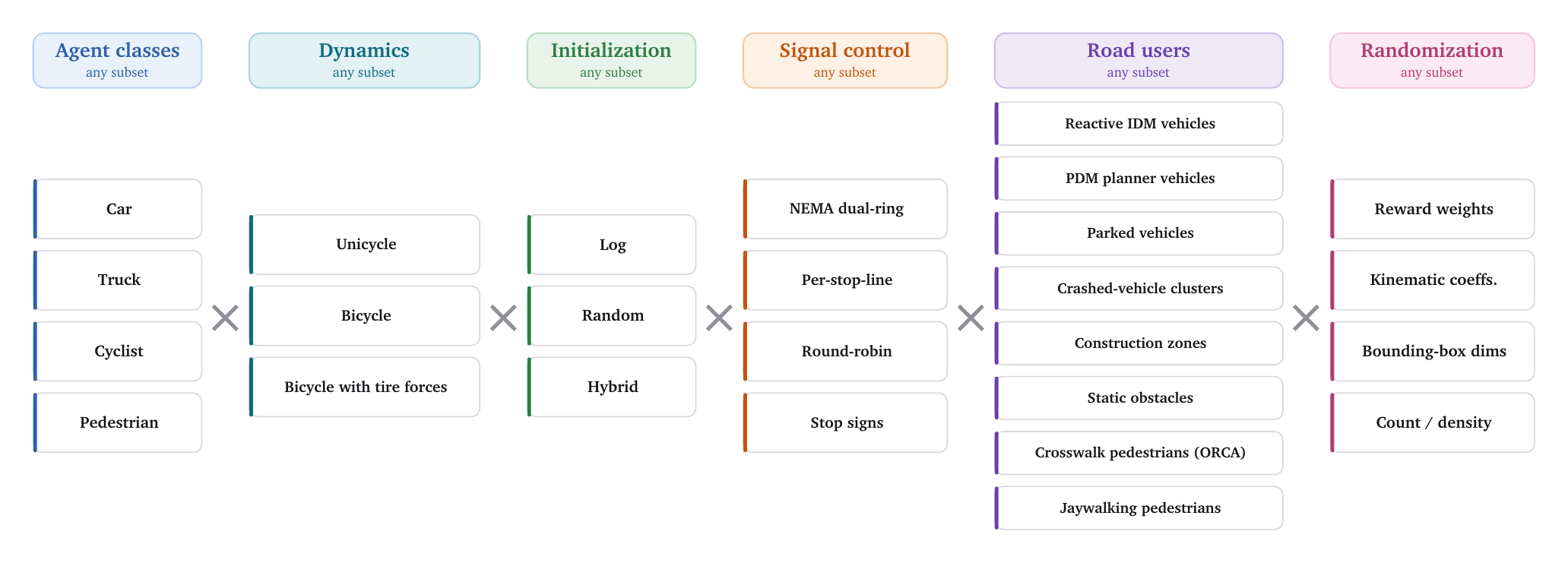}
\caption{\textbf{Composable configuration axes.} A scenario is composed by choosing a subset of options along each of six independent axes: agent classes, per-class dynamics, initialization, signal control, rule-based road users, and per-episode randomization. Because the axes compose freely, a single map seeds a combinatorially large space of scenarios.}
\label{fig:config_axes}
\end{figure}

Every subsystem in the engine is independently configurable, and the options compose. A scenario is assembled by choosing a subset of options along each axis in Figure~\ref{fig:config_axes}: agent classes, per-class dynamics, initialization, signal control, rule-based road users, and per-episode randomization. The axes compose independently, so the reachable scenario space grows combinatorially in the per-axis choices; the road-user axis, for instance, admits any subset of its eight generators (Section~\ref{subsec:npcs}). A new combination is a configuration change rather than a code change, which is what lets a single map seed the scenario distributions of Section~\ref{sec:scenario_generation}. The observation, reward, and kinematic subsystems expose the same switchable control, and the policy network resizes automatically to the enabled feature set (Section~\ref{sec:training}).

\paragraph{Per-class dynamics.}
Each agent class is simulated under its own dynamics model: pedestrians under a unicycle model, cyclists and cars under a bicycle model, and trucks under a bicycle model augmented with tire cornering forces. Cars are actuated at the jerk level for smooth, jerk-limited control, while the other classes are actuated by acceleration. A single shared policy controls all classes through per-type action masking (Section~\ref{sec:training}).

\subsection{Traffic Rule Enforcement}
\label{subsec:traffic-rules}

\methodname{} enforces the principal driving rules directly at the simulation layer: each violation drives a reward penalty (Section~\ref{sec:training}), enters the evaluation metrics, and can trigger a configurable consequence such as stopping or removing the offending agent.

\paragraph{Collisions and off-road.}
Collisions between agents and between an agent and an obstacle are resolved by a separating-axis test over oriented bounding boxes. Off-road excursions are caught by testing the agent's footprint against road-edge and sidewalk boundaries, with an elevation gate so that overpasses are ignored.

\paragraph{Lane direction.}
Lane-direction compliance is measured from the heading residual between the agent and its current lane, so reversed travel is penalized but never masked. Map handedness (left- versus right-hand drive) is fixed per map and carried in the lane geometry rather than in a separate check, so the same rule logic applies on both sides of the road.

\paragraph{Signals and stop signs.}
Signalized and sign-controlled intersections share a common geometric test: the agent's footprint is checked against per-stop-line regions, and entry and exit are tracked as an edge-triggered passage. A red-light violation registers when the agent's swept path crosses the stop bar on a red tick, not merely for waiting within the region on red, so a vehicle that halts at the line and proceeds on green is not penalized. Stop signs require the agent to hold below a speed threshold for a sampled dwell time before the line clears. Pedestrians are exempt from intersection rules, and cyclists are treated as vehicles. The signal state these checks consult comes from the controllers of Section~\ref{subsec:signals}.

\subsection{Data Pipeline}
\label{sec:data_pipeline}

An offline pipeline (the left branch of Figure~\ref{fig:system_architecture}) converts each dataset into a single common representation (inspired by ScenarioNet~\citep{li2023scenarionet}), decoupling source-specific parsing from the simulator. Separate converters ingest Waymo, nuPlan, and CARLA, and a builder compiles the shared representation into simulation-ready geometry: it identifies intersections, ties stop lines and traffic signals to their lanes, and resolves legal travel directions and speed limits. Intersections come from a source's own junction geometry where it provides one and are otherwise inferred from lane topology.

The builder validates each scene and emits it as a compact binary \emph{terrabin} file that the engine reads directly. This keeps the simulator independent of any single data source: supporting a new dataset requires only a new converter. At training time, scenarios are streamed and sampled so that each batch stays balanced across regions and scenario types without holding the full dataset in memory.

%% file: sections/scenario_generation.tex
\section{Procedural Scenario Generation}
\label{sec:scenario_generation}
A single real-world map is the seed for thousands of distinct training scenarios. \methodname{} procedurally generates a superset of the logged data rather than being bounded by it: recorded map geometry and trajectories anchor where roads run and where agents may begin, and a stack of composable randomization mechanisms expands each seed into the far broader distribution the policy actually trains on.

\subsection{Scene Initialization}
\label{subsec:scene-init}

\methodname{} supports three initialization modes: log, random, and hybrid. In log mode, agents are instantiated directly from logged dataset trajectories and inherit their recorded goals. In random mode, the mode used for our default training runs, agents are procedurally placed per class under rejection sampling, so that every agent starts collision-free and compliant with the traffic rules, on a lane from which a sufficiently distant goal is reachable (Appendix~\ref{app:random-placement}). The agent count is itself sampled from a configurable range whose upper bound may exceed the logged agent count, so scenes can be packed denser than any naturalistic recording. The hybrid mode mixes log and random at the granularity of whole environments through an independent per-environment draw, yielding a tunable blend across the batch rather than a within-scene mixture (Appendix~\ref{app:hybrid-init}).

\subsection{Goal Assignment}
\label{subsec:goals}

Procedurally initialized agents need navigation goals, which are sampled by a forward walk over the lane-topology graph up to a bounded arc length, filtered so that the goal lies ahead of the agent (Appendix~\ref{app:goal-assignment}). A configurable goal-dropout probability periodically hides the goal to encourage robust, non-goal-reliant behavior. When an agent reaches its goal, it either receives a freshly sampled goal further along the topology or halts in place, according to configuration.

\subsection{Actor Diversification}
\label{subsec:actor-div}

Beyond placement, the actors themselves are diversified. Each agent's bounding-box dimensions are sampled per episode from type-specific ranges, with vehicles further split across car, truck, and bus size classes (Appendix~\ref{app:random-placement}). Sampling dimensions independently of the source recording exposes the policy to a continuum of footprints, from compact cars to long buses and trucks, rather than the fixed sizes of any single dataset.

\begin{figure}[t]
\centering
\includegraphics[width=\textwidth]{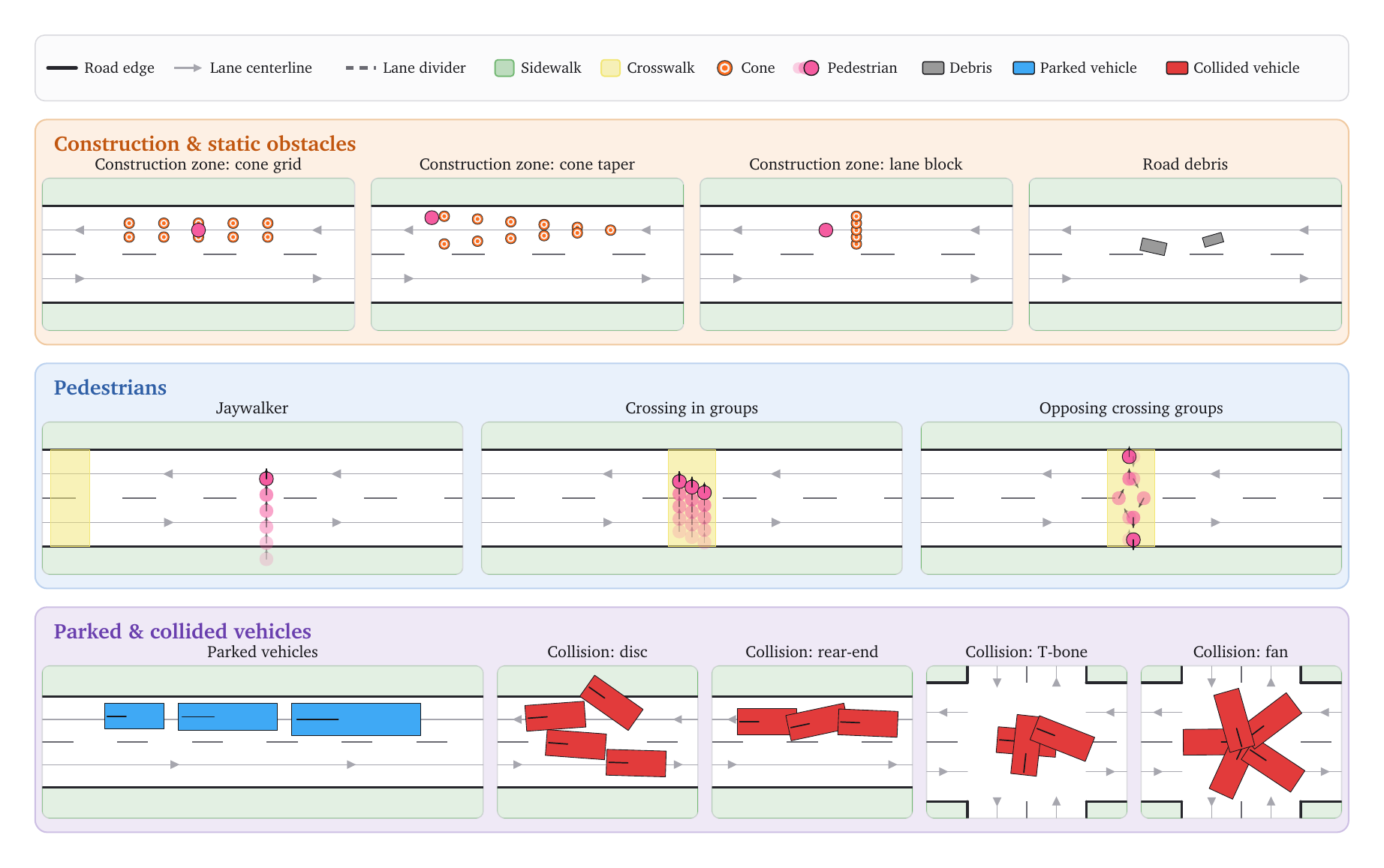}
\caption{\textbf{Rule-based road users.} Schematic of how \methodname{} procedurally generates its scripted road users, drawn on synthetic whole-lane frames with solid road edges, white dashed lane dividers, grey direction-carrying centerlines, and green sidewalks. The top row shows the construction and static-obstacle generators that close a lane (a tiled cone grid, a diagonal taper, and a perpendicular lane-block, each with a stationary worker, alongside random road debris); the middle row the pedestrian generators as faded temporal trails (a mid-block jaywalker, a one-direction pack, and an opposing bidirectional group bowing apart under ORCA avoidance); and the bottom row the vehicle generators (curb-anchored parked cars, buses, and trucks, and the four pre-arranged collided-vehicle clusters: uniform disc, rear-end chain, T-bone, and outward fan). The legend keys the road-surface markings and the entity glyphs.}
\label{fig:npc_gallery}
\end{figure}

\subsection{Rule-Based Non-Player Characters}
\label{subsec:npcs}

Alongside the learned agents, \methodname{} populates scenes with rule-based road users, or non-player characters (NPCs), which enrich interaction and, in self-play, help break the symmetry of a single shared policy training against copies of itself. The system offers several classes of these characters, and a scene draws a configurable mix of them that varies from episode to episode; the mixed presets used in our default training compose all of the spawned subsystems in one scene. Figure~\ref{fig:npc_gallery} shows how these scripted road users are procedurally generated.

\paragraph{Reactive vehicles.}
Reactive vehicles are driven either by an IDM~\citep{treiber2000idm} longitudinal controller paired with pure-pursuit steering, or by a closed-loop planner in the style of PDM-Closed~\citep{dauner2023pdm} that forward-simulates a grid of lateral-offset~$\times$~IDM-parameter proposals and selects the trajectory maximizing a weighted score of progress, time-to-collision, comfort, and lane-keeping. IDM vehicles sample among default, assertive, and cautious behavior modes that rescale the IDM gains, and may reroll that mode within an episode, producing heterogeneous and time-varying driving styles.

\paragraph{Static actors.}
Static actors stay fixed for the episode once placed. They include parked vehicles along curbs spanning the car, bus, and truck size classes, pre-arranged crashed-vehicle clusters in a uniform-disc, rear-end chain, T-bone, or outward-fan layout, construction zones built from tiled traffic cones in grid, taper, or lane-block closures with optional stationary workers, and isolated static obstacles offset from the lane center.

\paragraph{Pedestrians and cyclists.}
Pedestrians cross at crosswalks under a reciprocal collision-avoidance controller (ORCA), arriving on a Poisson schedule sampled per crosswalk as single crossers, packs, bidirectional flows, or staggered sequences, and also jaywalk mid-block away from marked crossings, walking perpendicular to the road until they clear the far edge. Cyclists have no scripted controller; \methodname{} populates them as NPCs only through log replay, though the learned policy still controls them as agents.

\subsection{Signal Control}
\label{subsec:signals}

The intersections in a scene are driven by one of three configurable traffic-light controllers; detection of the resulting violations is handled by the engine and described in Section~\ref{subsec:traffic-rules}. A \emph{NEMA} controller runs the standard dual-ring, eight-phase concurrency plan~\citep{nema2003ts2}: each ring advances its phases through green, yellow, and all-red intervals, and the two rings cross barriers together so that conflicting movements never run at once. The \emph{Christmas controller}, used in our default training composition, instead cycles each stop line independently through red, green, and yellow with dwell times drawn from a LogNormal distribution, so signals across a map stay uncoordinated. A \emph{round-robin} controller is the simplest: one approach leg holds green at a time, and the green window rotates leg by leg in canonical NEMA order. Two further mechanisms inject per-scene variety: the initial phase of each intersection is randomized at reset, and protected left turns are flipped to permissive by an independent per-intersection draw, so the same map presents different signal timing and turn permissions across episodes.

%% file: sections/training.tex
\section{A Robust Training Recipe}
\label{sec:training}

\methodname{} is trained with a self-play PPO recipe~\citep{schulman2017ppo} built on PufferLib~\citep{suarez2024pufferlib} and deliberately co-designed with the fast object-level simulator of Section~\ref{sec:system}. Training is \emph{tabula rasa}: the policy starts from randomly initialized weights and learns from reinforcement signal alone, with zero human demonstrations: no imitation loss and no dependence on logged trajectories. Logged data enters only as map geometry and as the optional starting distribution of Section~\ref{sec:scenario_generation}, so the default composition initializes agents randomly, and log and hybrid initialization are available when logged initial states are preferred but never required. The central observation behind the recipe is that when rollouts are cheap, the right currency is \emph{compute} efficiency rather than sample efficiency: it is worth discarding low-value samples and spending the freed compute on aggressive off-policy corrections, value normalization, and a heterogeneous agent population.

\subsection{Compute-Efficient RL}
\label{subsec:compute-efficient}

\paragraph{Saliency-prioritized sampling.}
Rather than sweeping every collected transition, \methodname{} draws a fixed number of minibatches by sampling trajectory segments \emph{with replacement} in proportion to their priority $p_i \propto \big(\sum_t |\hat{A}_{i,t}|\big)^{\alpha}$, following the prioritized-replay principle~\citep{schaul2016per}. Because the simulator is fast, this is a favorable trade: high-advantage segments are oversampled while near-zero-advantage segments are simply skipped, yielding more gradient signal per unit of compute at the cost of revisiting fewer unique samples. To correct the induced bias we apply importance-sampling weights $w_i = (N p_i)^{-\beta}$ with $\beta$ annealed upward over training.
\paragraph{Compact feed-forward policy.}
The default \methodname{} policy is a compact multilayer perceptron (MLP) network of approximately $3.5$M parameters, in the style of \citet{cusumano-towner2025gigaflow} but with a simpler observation set: a single road-geometry resolution rather than separate coarse and fine map views, a single relative-goal target rather than a routed distance field with intermediate waypoints, and a smaller partner neighborhood (Appendix~\ref{app:observations}). Per-group encoders (an ego MLP and permutation-invariant Deep Sets encoders~\citep{zaheer2017deepsets} for the road, partner, and traffic sets) project each group to an embedding, which is concatenated and passed through a three-layer, $1024$-unit MLP shared by the actor and value heads. The value bootstrap is handled separately and is preserved across truncations (next paragraph).

\paragraph{Markov decision process design.}
Each agent observes its ego state (speed, dimensions, heading, steering, acceleration, a relative goal, and a one-hot agent type), up to $20$ partner agents within $50$\,m, and up to $200$ nearby road segments, with the partner and road sets consumed by the Deep Sets encoders. Vehicles act in a discrete control space, in either acceleration or jerk. We train two kinds of policy on this substrate: a \emph{planner} that controls vehicles alone, and a heterogeneous \emph{sim agent} (Section~\ref{subsec:wosac}) that drives all three classes through separate per-type action heads over their own dynamics models: jerk for vehicles, a unicycle grid for pedestrians, and a compact bicycle grid for cyclists (Appendix~\ref{app:hyperparameters}). The reward follows the Gigaflow shaping~\citep{cusumano-towner2025gigaflow}, combining a goal bonus with collision, off-road, comfort, lane-alignment, lane-centering, velocity, and reverse-driving penalties; the full specification and default weights are given in Appendix~\ref{app:rewards}, and observation feature groups in Appendix~\ref{app:observations}.

\paragraph{Value estimation and learning dynamics.}
Advantages are computed by generalized advantage estimation (GAE)~\citep{schulman2016gae}, augmented with three stability mechanisms.
\emph{V-trace off-policy corrections}~\citep{espeholt2018impala} clip the temporal-difference and trace terms; because the policy changes between rollout collection and the gradient update, the clipped importance ratio from each minibatch is stored and used to recompute advantages at the start of the next epoch. Setting both clips to infinity recovers standard GAE. True terminals zero the value bootstrap while truncations preserve a saved $V(s_\text{final})$ from the pre-reset observation; both cut the $\lambda$ trace, so the policy is never penalized for surviving near a scenario boundary, a documented bias in time-limited RL~\citep{pardo2018timelimits}.
\emph{PopArt value normalization}~\citep{vanhasselt2016popart} addresses return magnitudes that vary by orders of magnitude across scenarios: running mean $\mu$ and standard deviation $\sigma$ are updated by an exponential moving average each epoch, and the value head is analytically rescaled so that its denormalized predictions are preserved across statistic updates. Value targets are computed in normalized space while advantages enter GAE on the true return scale.
The three mechanisms (priority sampling, V-trace, and PopArt) address complementary aspects of the off-policy self-play optimization; Appendix~\ref{app:hyperparameters} lists the values of every optimization hyperparameter.

\paragraph{Domain randomization over rewards and dynamics.}
The training environment is randomized along two axes that the policy observes and must adapt to. The reward function is a vector of weighted terms (goal, collision, off-road, comfort, lane-alignment, lane-centering, velocity, and reverse-driving, among others), each of which can be disabled or randomized per agent within a range sampled at episode reset, following the heterogeneous-agent formulation of Gigaflow~\citep{cusumano-towner2025gigaflow}. The same randomization applies to the four multiplicative kinematic-scaling coefficients $(c_\text{throttle}, c_\text{steer}, c_\text{acc}, c_\text{vel})$ that govern each agent's acceleration, steering, and speed limits. Both the sampled reward weights and the kinematic coefficients enter the ego observation (reward- and kinematic-conditioned observations), so the policy adapts to the current regime online rather than memorizing a single one, and its behavior can be steered at inference by adjusting the weight vector without retraining (Appendices~\ref{app:rewards}, \ref{app:kin-random}, and~\ref{app:observations}). Our default training composition randomizes both axes.

\paragraph{Goal dropout.}
\emph{Goal dropout} keeps the policy from over-relying on its goal input. It marks a per-type fraction of the policy-controlled agents at each reset ($0.3$ of vehicles by default) and hides the goal of a marked agent by zeroing its relative-goal observation, while a separate visible flag tells the network the goal is masked rather than positioned at the origin. The marked agent must continue driving sensibly with its goal hidden, following the road structure until it leaves the map or the episode ends. \methodname{} also exposes an optional observation-noise model, off by default, that adds clipped Gaussian perturbations and slot dropout to the ego, partner, and road features for perception-robustness experiments.

\paragraph{Breaking self-play asymmetry with population play.}
A single shared policy controls all learning agents, so self-play pits the policy against copies of itself; left unchecked this invites degenerate, perfectly symmetric equilibria. \methodname{} breaks the symmetry by populating each scene with a heterogeneous \emph{population} of controllers rather than a league of frozen checkpoints. Two ingredients do the work. First, the per-agent kinematic and reward domain randomization described above makes co-trained policy agents behave heterogeneously even under one network. Second, scenes are populated with the rule-based road users of Section~\ref{subsec:npcs}: reactive IDM controllers~\citep{treiber2000idm} and closed-loop PDM planners~\citep{dauner2023pdm}, alongside parked vehicles, crashed-vehicle clusters, construction zones, and crossing or jaywalking pedestrians, in a configurable mix that varies from scene to scene (Figure~\ref{fig:npc_gallery}). The learned policy must therefore remain robust to interacting with agents whose behavior it does not control, which is precisely the situation it faces in mixed-control deployment and evaluation.

\subsection{Distributed Training}
\label{subsec:distributed}
\methodname{} scales across GPUs and nodes by data parallelism: each rank runs its own simulation environments and computes local gradients over its own rollouts, and the gradients are all-reduced across ranks into a single global mean update.

The subtlety in distributed self-play is not the gradient all-reduce but keeping the recipe's \emph{normalization statistics} globally consistent, since each stability mechanism maintains running statistics that are meaningless if computed per-rank. \methodname{} synchronizes these across ranks so that advantage normalization, PopArt value rescaling, and joint priority sampling all operate on a single global scale rather than on any one rank's shard.

%% file: sections/experiments.tex
\section{Experiments \& Results}
\label{sec:experiments}

\subsection{Simulation Throughput}
\label{subsec:throughput}

\begin{figure}[t]
\centering
\includegraphics[width=\textwidth]{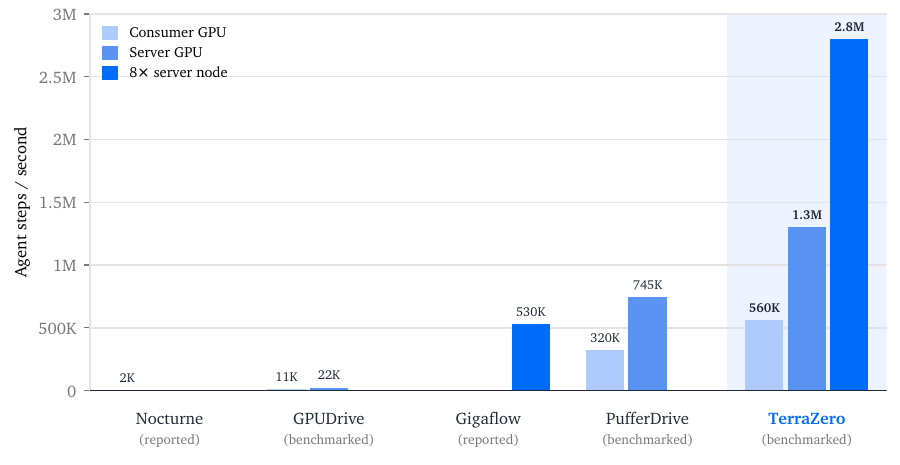}
\caption{\textbf{Training throughput.} Agent steps per second across hardware configurations. \methodname{} achieves competitive throughput on a single consumer GPU and the highest throughput on server hardware, while retaining full scenario fidelity including traffic signals, heterogeneous agents, and reactive traffic.}
\label{fig:throughput}
\end{figure}

A key claim of \methodname{} is high throughput \emph{without} sacrificing scenario fidelity. Figure~\ref{fig:throughput} shows agent-steps-per-second scaling across hardware, and the head-to-head comparison against existing simulators is summarized in Table~\ref{tab:simulator_comparison} (Section~\ref{sec:related_work}). Measured on a representative training job, \methodname{} sustains $560$K agent-steps per second on a single consumer GPU, $1.3$M on a single server-grade GPU, and $2.8$M on an $8$-GPU server node, to our knowledge significantly faster than any existing driving simulator. It reaches these rates while supporting three agent classes, full traffic-signal state machines, and reactive traffic, a fidelity that the single-agent backends it outpaces do not provide.

\subsection{Policy Training Setup}
\label{subsec:experimental-setup}

\paragraph{Training data.}
We source driving scenarios from three datasets, each converted to the terrabin format of Section~\ref{sec:data_pipeline}, and we match each dataset to the experiments it supports. nuPlan~\citep{caesar2021nuplan} supplies the map geometry for the planner evaluated on val14 (Section~\ref{subsec:val14}) and InterPlan (Section~\ref{subsec:interplan}). WOMD~\citep{ettinger2021womd} supplies the map geometry for the sim agent evaluated on WOSAC (Section~\ref{subsec:wosac}), which trains on the full WOMD training split and is scored zero-shot on the WOSAC validation sets described there. The transfer study (Section~\ref{subsec:generalization}) draws on all three sources, and their relative scale sets its context.

WOMD is the largest of the three: our direct converter yields about $576$K scenarios, of which $487$K form the training split. nuPlan contributes about $262$K scenarios across its four cities, distributed unevenly: Las Vegas is by far the largest and Boston the smallest, with Pittsburgh and Singapore in between. CARLA~\citep{dosovitskiy2017carla} adds only five synthetic towns as map geometry. These sources span a broad distribution of intersections, highway merges, roundabouts, and urban corridors while differing in scale by orders of magnitude, from five hand-built CARLA maps to nearly half a million WOMD scenarios.

\paragraph{Policy and training.}
Unless otherwise noted, experiments use the compact multilayer-perceptron policy of Section~\ref{sec:training} with jerk-based vehicle actions, trained from scratch with zero human demonstrations on nuPlan map geometry with PPO~\citep{schulman2017ppo} and GAE~\citep{schulman2016gae}, V-trace off-policy corrections~\citep{espeholt2018impala}, and PopArt value normalization~\citep{vanhasselt2016popart}, in bf16 mixed precision distributed across 16 NVIDIA A100 80GB GPUs. We apply kinematic, reward, and agent-density domain randomization throughout training, and the configuration of this planner checkpoint, shared by the val14 and InterPlan evaluations, is summarized in Appendix~\ref{app:hyperparameters}. The heterogeneous sim agent of Section~\ref{subsec:wosac} is the exception: it uses separate per-type action heads, about $6.7$M parameters, and trains on 32 GPUs.

\subsection{Planner Benchmarks}
\label{subsec:planning}

We evaluate \methodname{} as a \emph{planner}, an ego driving policy, on two closed-loop ego planning benchmarks. The standard val14 split (Section~\ref{subsec:val14}), run through the PlanTF evaluation suite~\citep{cheng2024plantf}, draws 14 common nuPlan scenario types from logged validation scenes in balanced proportion, measuring in-distribution urban driving. The InterPlan interactive long-tail suite (Section~\ref{subsec:interplan})~\citep{hallgarten2024interplan} rewrites logged nuPlan scenes into the rare, interaction-heavy situations that val14 seldom contains, such as construction zones, accident sites, and jaywalking pedestrians. The two benchmarks score the policy with the official nuPlan closed-loop reactive score and execute its actions through the benchmark's standard nuPlan Linear-Quadratic Regulator (LQR) controller, the same low-level tracker every other planner runs under, so no controller-side change contributes to either score. At evaluation \methodname{} conditions the policy for each benchmark through the reward- and kinematic-conditioned observations of Section~\ref{sec:training}, with no retraining, and aligns the eval-time goal and route sampling with the lane-level goals the policy saw during training (Section~\ref{subsec:goals}); the benchmarks' own scoring stays unchanged.

\subsubsection{nuPlan val14}
\label{subsec:val14}

Our primary ego-policy evaluation uses the nuPlan val14 closed-loop reactive benchmark~\citep{caesar2021nuplan}, run through the PlanTF evaluation suite~\citep{cheng2024plantf} and scored with the official nuPlan closed-loop reactive score. Each rollout runs the ego policy in closed loop against reactive traffic, and the score aggregates no-at-fault collision, time-to-collision (TTC), drivable-area, driving-direction, and speed-limit compliance, ego progress, making progress, and comfort into a single weighted value in $[0,100]$, reported in Table~\ref{tab:val14} alongside its component terms. The comparison set spans the four planner families that populate the val14 leaderboard (rule-based, imitation, hybrid rule-plus-learned, and reinforcement learning), placing \methodname{} against both the hand-engineered planners that top the benchmark and the learned policies nearest its own setting; the closest of these is Gigaflow~\citep{cusumano-towner2025gigaflow}, the one comparable large-scale GPU self-play system. \methodname{} is, to our knowledge, the first policy trained exclusively via large-scale self-play to outscore the dedicated nuPlan planners that top val14.

\begin{table}[t]
\centering
\caption{Driving policy evaluation on nuPlan val14 (closed-loop reactive). All metrics are higher-is-better. Type: Rule = rule-based; IL = imitation learning; Hybr. = hybrid rule+learned; RL = reinforcement learning; Replay = log replay with tracking controller. Asterisk (*) denotes author re-trained variants; the dagger ($\dagger$) marks the history-free G2DP variant with rule-based post-hoc refinement (+ref.); dashes mark component scores the source paper does not report. SPDM appears at two proposal budgets $N_p$, its val14-best ($N_p{=}15$) and its InterPlan-best ($N_p{=}60$; Section~\ref{subsec:interplan}). \textbf{Bold} indicates best; \underline{underline} indicates second best. The Log Replay reference row is excluded from the best and second-best marking.}
\label{tab:val14}
\scriptsize
\setlength{\tabcolsep}{3pt}
\resizebox{\textwidth}{!}{%
\begin{tabular}{@{}llccccccccc@{}}
\toprule
Planner & Type & Score $\uparrow$ & Ego Prog. $\uparrow$ & No AF-Coll. $\uparrow$ & Comfort $\uparrow$ & TTC $\uparrow$ & Drv.~Dir. $\uparrow$ & Speed~Lim. $\uparrow$ & Drv.~Area $\uparrow$ & Making~Prog. $\uparrow$ \\
\midrule
Log Replay (iLQR)                          & Replay  & 82.05 & 99.32 & 85.69 & 98.93 & 80.77 & 99.28 & 96.46 & 99.55 & 100.00 \\
\midrule
IDM~\citep{treiber2000idm}                 & Rule    & 77.33 & 85.20 & 89.22 & 93.11 & 81.57 & 99.24 & 97.20 & 93.02 & 96.33 \\
PDM-Closed~\citep{dauner2023pdm}           & Rule    & 92.13 & 90.26 & 97.90 & 94.72 & 93.83 & \textbf{99.96} & \underline{99.83} & 99.46 & \textbf{99.11} \\
SPDM ($N_p{=}15$)~\citep{distelzweig2026spdm} & Rule  & 92.28 & --    & --    & --    & --    & --    & --    & --    & --    \\
SPDM ($N_p{=}60$)~\citep{distelzweig2026spdm} & Rule  & 91.60 & --    & --    & --    & --    & --    & --    & --    & --    \\
PLUTO~\citep{cheng2024pluto}               & Hybr.   & 89.66 & 86.11 & 97.09 & 91.86 & \underline{94.10} & \underline{99.87} & 98.86 & 99.28 & 98.57 \\
FlowDrive w/ guidance + PDM scoring~\citep{wang2025flowdrive} & Hybr. & 92.96 & --    & --    & --    & --    & --    & --    & --    & --    \\
G2DP$^\dagger$ +ref.~\citep{yu2026g2dp}    & Hybr.   & 92.92 & --    & --    & --    & --    & --    & --    & --    & --    \\
Urban Driver~\citep{scheel2022urbandriver} & IL      & 53.05 & 87.17 & 72.45 & \textbf{99.28} & 66.37 & 96.87 & 81.66 & 83.36 & 94.28 \\
PlanTF~\citep{cheng2024plantf}             & IL      & 76.14 & 77.21 & 95.21 & 93.56 & 90.61 & 99.33 & 98.50 & 96.33 & 89.53 \\
Diffusion Planner*~\citep{zheng2025diffplanner} & IL & 82.73 & 85.88 & 93.07 & 89.45 & 88.19 & 99.82 & 98.29 & 97.85 & 96.42 \\
Flow Planner~\citep{tan2025flowplanner}    & IL      & 83.31 & --    & --    & --    & --    & --    & --    & --    & --    \\
G2DP~\citep{yu2026g2dp}                     & IL      & 83.91 & --    & --    & --    & --    & --    & --    & --    & --    \\
FlowDrive~\citep{wang2025flowdrive}        & IL      & 85.37 & --    & --    & --    & --    & --    & --    & --    & --    \\
Gigaflow~\citep{cusumano-towner2025gigaflow} & RL    & \underline{93.8}  & \textbf{93.6}  & \underline{98.4}  & 96.4  & 93.8  & 99.6  & \textbf{99.9}  & \underline{99.7}  & 99.0  \\
CaRL~\citep{jaeger2025carl}                  & RL    & 90.60 & 91.30 & 97.05 & 88.55 & 92.31 & 99.15 & 99.36 & \textbf{99.91} & \textbf{99.11} \\
PlannerRFT~\citep{li2026plannerrft}          & RL    & 84.46 & --    & --    & --    & --    & --    & --    & --    & --    \\
\midrule
\methodname{} (Ours)                       & RL      & \textbf{94.19} & \underline{92.91} & \textbf{99.02} & \underline{97.94} & \textbf{95.62} & 99.51 & 99.27 & 99.37 & \underline{99.02} \\
\bottomrule
\end{tabular}%
}
\end{table}

\methodname{} scores $94.19$ on val14, the highest score among the planners we compare against. It leads the reinforcement-learning family, ahead of Gigaflow~\citep{cusumano-towner2025gigaflow} at $93.8$, CaRL~\citep{jaeger2025carl} at $90.60$, and PlannerRFT~\citep{li2026plannerrft}, a diffusion planner refined by reinforcement fine-tuning that reaches $84.46$. The strongest rule-based and hybrid entries, which hand-craft trajectories or graft a rule-based scorer onto a learned model, trail as well: the guidance-plus-PDM variant of FlowDrive~\citep{wang2025flowdrive} at $92.96$, the refinement-augmented G2DP$^\dagger$~\citep{yu2026g2dp} at $92.92$, SPDM at its val14-tuned 15-proposal setting~\citep{distelzweig2026spdm} at $92.28$, and PDM-Closed~\citep{dauner2023pdm} at $92.13$. That SPDM setting is the very one that gives back most of its long-tail performance on InterPlan (Section~\ref{subsec:interplan}). \methodname{} reaches this with a markedly leaner observation set than Gigaflow: a single road-geometry resolution rather than separate coarse and fine map views, a single relative goal rather than a routed distance field with intermediate waypoints, and a smaller partner neighborhood (Section~\ref{sec:training}). Because the policy depends on none of these engineered map features, it stays usable where the underlying lane and road topology is imperfect or broken, the regime that richer feature stacks handle poorly. The safety terms carry the score: \methodname{} posts the best no-at-fault-collision ($99.02$) and time-to-collision ($95.62$) figures in the table, the safest of the compared planners on both, and the second-best comfort ($97.94$). Ego progress is the one component where a leader still edges it, Gigaflow at $93.6$ against $92.91$, so the residual headroom sits in progress rather than in safety.

This result covers only the in-distribution scenario types that val14 samples. Gigaflow trains on synthetic maps under a single standard-driving distribution and reports no long-tail benchmark, whereas \methodname{} trains on both the fat body of routine driving and the procedurally generated long tail of Section~\ref{subsec:npcs}. The next benchmark tests whether that long-tail training pays off where val14 is silent.

\subsubsection{InterPlan}
\label{subsec:interplan}

We next evaluate \methodname{} on InterPlan~\citep{hallgarten2024interplan}, an interactive closed-loop benchmark that stress-tests planning in the long-tail, out-of-distribution situations that val14 rarely contains. InterPlan modifies logged nuPlan scenes into eight interactive scenario families: construction zones, accident sites, jaywalking pedestrians, nudging around a stopped vehicle, overtaking with oncoming traffic, and lane changes at low, medium, and high traffic density. Each planner drives in closed loop against reactive agents and is scored by the nuPlan closed-loop reactive score, which aggregates safety, progress, comfort, and compliance into a single value in $[0,100]$. We report on the official 80-scenario split so that every method is scored on the same scenarios, and we report \methodname{} on the larger 335-scenario set as a broader reference. The planner reported here and in Section~\ref{subsec:val14} is a single checkpoint trained with one configuration, summarized in Appendix~\ref{app:hyperparameters}.

\begin{table}[t]
\centering
\caption{Driving policy evaluation on the InterPlan interactive long-tail benchmark, scored by the closed-loop reactive nuPlan score in $[0,100]$ (higher is better). The InterPlan column reports the official 80-scenario split and the Full-Scale InterPlan column the larger 335-scenario set; dashes mark splits a method does not report. Type: Rule = rule-based; IL = imitation learning; Hybr. = hybrid, a learned or language model combined with a rule-based planner, including the LLM planners whose low-level controller is PDM-Closed; RL = reinforcement learning. Scores are from \citet{hallgarten2024interplan}, except Diffusion Planner, PPO, and PDM+PPO, which are from \citet{behaviorbenchmark2026} on the same reactive benchmark, and SPDM, FlowDrive, Flow Planner, and G2DP, which are from their own papers. \textbf{Bold} is best and \underline{underline} second best on the InterPlan split.}
\label{tab:interplan}
\small
\begin{tabular}{@{}llcc@{}}
\toprule
Planner & Type & InterPlan $\uparrow$ & Full-Scale InterPlan $\uparrow$ \\
\midrule
IDM~\citep{treiber2000idm}                          & Rule  & 31    & -- \\
IDM+MOBIL~\citep{treiber2000idm, kesting2007mobil}  & Rule  & 31    & -- \\
PDM-Closed~\citep{dauner2023pdm}                    & Rule  & 42    & -- \\
SPDM ($N_p{=}15$)~\citep{distelzweig2026spdm}       & Rule  & 42.00 & -- \\
SPDM ($N_p{=}60$)~\citep{distelzweig2026spdm}       & Rule  & \underline{63.66} & -- \\
DTPP~\citep{huang2024dtpp}                          & Hybr. & 25    & -- \\
HybridLLMPlanner (GPT-3.5)~\citep{hallgarten2024interplan}   & Hybr. & 40    & -- \\
HybridLLMPlanner (Llama-13B)~\citep{hallgarten2024interplan} & Hybr. & 48    & -- \\
HybridLLMPlanner (Llama-7B)~\citep{hallgarten2024interplan}  & Hybr. & 53    & -- \\
PDM+PPO~\citep{behaviorbenchmark2026}               & Hybr. & 43.2  & -- \\
FlowDrive w/ guidance + PDM scoring~\citep{wang2025flowdrive} & Hybr. & 44.05 & -- \\
Urban Driver~\citep{scheel2022urbandriver}          & IL    & 4     & -- \\
GC-PGP~\citep{hallgarten2023gcpgp}                  & IL    & 10    & -- \\
GameFormer~\citep{huang2023gameformer}              & IL    & 11    & -- \\
PDM-Open~\citep{dauner2023pdm}                      & IL    & 25    & -- \\
Diffusion Planner~\citep{zheng2025diffplanner}      & IL    & 25.8  & -- \\
FlowDrive~\citep{wang2025flowdrive}                 & IL    & 36.96 & -- \\
Flow Planner~\citep{tan2025flowplanner}             & IL    & --    & 61.82 \\
G2DP~\citep{yu2026g2dp}                              & IL    & --    & 61.74 \\
PPO~\citep{behaviorbenchmark2026}                   & RL    & 42.1  & -- \\
\midrule
\textbf{\methodname{} (Ours)}                       & RL    & \textbf{70.87} & \textbf{71.31} \\
\bottomrule
\end{tabular}
\end{table}

\methodname{} attains a score of $70.87$ on the 80-scenario split, ahead of every prior planner; the closest is SPDM, a proposal-enriched PDM variant, at $63.66$~\citep{distelzweig2026spdm}, and the strongest LLM-based method, the 7B-parameter LLaMA HybridLLMPlanner, reaches $53$~\citep{hallgarten2024interplan}. SPDM reaches $63.66$ only at the widest proposal budget it evaluates, 60 candidates per step, and its own sweep reveals a benchmark-specific tuning tax, which is why we tabulate its two endpoints as separate rows in Tables~\ref{tab:val14} and~\ref{tab:interplan}: the 15-proposal setting peaks on val14 at $92.28$ but scores only $42.00$ on InterPlan, while the 60-proposal setting that reaches $63.66$ on InterPlan gives back val14, down to $91.60$. No single SPDM configuration is strong on both benchmarks. The base PDM-Closed planner shows the same split, strong on val14 ($92.13$) yet weak on InterPlan ($42$), a tradeoff characteristic of rule-based planners, whose hand-designed proposal and scoring logic is tuned to one regime and does not carry to the long tail without cost elsewhere. One \methodname{} checkpoint tops both benchmarks, exceeding SPDM's best val14 ($94.19$ versus $92.28$) and its best InterPlan ($70.87$ versus $63.66$). \methodname{} reaches this with the compact multilayer-perceptron policy of Section~\ref{sec:training}, over three orders of magnitude smaller than a 7B-parameter language model, which points to procedural long-tail scenario generation rather than model scale as the source of the gain: the reported policy trains against a dense, procedurally generated population of construction cones, static obstacles, crashed-vehicle clusters, crossing and jaywalking pedestrians, and reactive IDM traffic (Section~\ref{subsec:npcs}), so the long-tail families InterPlan probes resemble scenes the policy already practices.

The comparison also separates learned control from rule-based control. The strongest prior entries lean on a hand-crafted planner: the highest prior score is rule-based (SPDM at $63.66$), and the competitive hybrids graft a rule-based planner or a language model onto a learned model, where HybridLLMPlanner pairs a language model with a PDM-Closed fallback, PDM+PPO~\citep{behaviorbenchmark2026} combines PDM with a learned policy, and FlowDrive climbs from $36.96$ to $44.05$ when its trajectories are reranked by the PDM scorer~\citep{wang2025flowdrive}. Purely learned planners that carry no such planner score far lower (Urban Driver $4$, GameFormer $11$, Diffusion Planner $25.8$, FlowDrive $36.96$, PPO $42.1$). \methodname{} uses no rule-based planner and no language model at inference, so the result comes from the learned policy alone. It is, to our knowledge, the first fully learned policy to reach the top of this interactive long-tail benchmark.

The score is bounded by progress rather than by safety. On the 335-scenario set the route-progress term sits at $67.1$, the lowest of the components, with driving direction next at $88.8$, while time-to-collision ($93.4$), no-fault collision ($95.5$), and drivable-area compliance ($97.3$) all stay high. The policy keeps agents on the road and largely collision-free even in these rare, deliberately adversarial scenes, and the remaining headroom is in making forward progress through them. The same policy scores $71.31$ on that set under the identical training configuration, ahead of the strongest prior learned planners reporting on it, Flow Planner at $61.82$~\citep{tan2025flowplanner} and G2DP at $61.74$~\citep{yu2026g2dp}, which confirms that the result is not an artifact of the smaller official split.

\subsection{Sim Agent Benchmark}
\label{subsec:wosac}

We next evaluate \methodname{} as a \emph{sim agent}, a controllable traffic actor whose behavior must match the statistical distribution of real human driving. WOSAC characterizes each scenario as a 9.1\,s WOMD recording at 10\,Hz, uses the first 1.1\,s as initial context, and scores how well a method reproduces the remaining 8\,s for up to 128 agents under 32 stochastic closed-loop joint rollouts, combined into a realism meta-metric over kinematic, interactive, and map-based likelihood scores~\citep{montali2023wosac}. We report both editions on the Waymo validation set, evaluated zero-shot, and match the scenario selection of the competing methods in each edition. The 2023 edition uses the full validation split after removing scenarios above the 128-agent challenge cap, as in Gigaflow~\citep{cusumano-towner2025gigaflow}. The 2024 edition uses the filtered validation subset of CAT-K~\citep{zhang2025catk} and SPACeR~\citep{chang2026spacer}, built from the released CAT-K validation manifest under the same 128-agent cap. We use the 2023 edition to compare against Gigaflow, the closest demonstration-free self-play system to ours, and the 2024 edition to compare against more recent self-play methods.

\paragraph{Implementation details.}
The \methodname{} sim agent is a heterogeneous self-play policy that controls vehicles, pedestrians, and cyclists jointly through a single shared network with separate per-type action heads over jerk (vehicles), unicycle (pedestrians), and compact-bicycle (cyclists) dynamics, trained without demonstration trajectories on 32 A100 GPUs (Appendix~\ref{app:hyperparameters}). The primary checkpoint trains on Waymo map geometry, and the zero-shot transfer checkpoint trains on nuPlan map geometry. Each checkpoint produces both the vehicle and vulnerable-road-user rollouts reported below.

\paragraph{Baselines.}
We compare against representative imitation and self-play approaches, and we draw the 2024 peer numbers from the SPACeR paper~\citep{chang2026spacer}, whose benchmark matches ours. The imitation family comprises SMART~\citep{wu2024smart} and SMART fine-tuned with CAT-K~\citep{zhang2025catk}, two tokenized models trained directly on WOMD demonstrations and reported as reference upper bounds. The self-play family comprises decentralized PPO trained on the task reward alone~\citep{kazemkhani2025gpudrive}, Human-Regularized PPO (HR-PPO)~\citep{cornelisse2024hrppo}, and SPACeR~\citep{chang2026spacer}. These methods differ in how much logged data they require: HR-PPO regularizes its policy toward a behavior-cloning reference and SPACeR anchors self-play to a pretrained tokenized reference model through Kullback--Leibler (KL) alignment, so each depends on a separate reference policy trained on logged data even though neither consumes demonstrations in the self-play loss, whereas PPO and \methodname{} use no logged human data of any kind. PPO, HR-PPO, and SPACeR also initialize agents and goals from logged trajectories, while \methodname{} uses procedural random initialization. The 2023 edition adds the published Gigaflow results~\citep{cusumano-towner2025gigaflow} alongside the random-agent, linear-extrapolation, and stationary lower bounds reported by the challenge organizers~\citep{montali2023wosac}.

\begin{table}[t]
\centering
\caption{WOSAC 2023 sim agent evaluation against demonstration-free self-play baselines, scored on the full Waymo validation split. All metrics are higher-is-better. \methodname{} raises overall realism while jointly controlling vehicles, pedestrians, and cyclists in closed loop, where Gigaflow drives pedestrians with a scripted controller. The shaded row is the logged expert reference; \textbf{bold} marks the best demonstration-free entry per column and \underline{underline} the second best.}
\label{tab:wosac2023}
\footnotesize
\setlength{\tabcolsep}{0.9pt}
\resizebox{\textwidth}{!}{%
\begin{tabular}{@{}lcccccccccc@{}}
\toprule
Method
 & \makecell{Realism\\Meta $\uparrow$}
 & \makecell{Lin.\\Speed $\uparrow$}
 & \makecell{Lin.\\Accel.\ $\uparrow$}
 & \makecell{Ang.\\Speed $\uparrow$}
 & \makecell{Ang.\\Accel.\ $\uparrow$}
 & \makecell{Dist.\\Obj.\ $\uparrow$}
 & \makecell{TTC\\$\uparrow$}
 & \makecell{Dist.\\Rd.\ Edge $\uparrow$}
 & \makecell{Off-\\road $\uparrow$}
 & \makecell{Coll.\\$\uparrow$} \\
\midrule
\rowcolor[gray]{0.93} Expert Demonstration~\citep{montali2023wosac}
 & 0.722 & 0.56 & 0.33 & 0.56 & 0.49 & 0.49 & 0.88 & 0.71 & 1.00 & 1.00 \\
\midrule
Random Agent~\citep{montali2023wosac}
 & 0.155 & 0.00 & 0.04 & 0.07 & 0.12 & 0.00 & 0.73 & 0.18 & 0.29 & 0.00 \\
Linear Extrapolation~\citep{montali2023wosac}
 & 0.324 & 0.16 & 0.12 & 0.02 & 0.04 & 0.25 & 0.78 & 0.50 & 0.46 & 0.41 \\
Stationary Policy~\citep{cusumano-towner2025gigaflow}
 & 0.501 & 0.04 & 0.06 & 0.50 & 0.32 & 0.06 & 0.74 & 0.24 & 0.86 & 0.95 \\
Gigaflow (zero-shot)~\citep{cusumano-towner2025gigaflow}
 & 0.619 & \underline{0.26} & \textbf{0.25} & \underline{0.51} & \textbf{0.48} & \textbf{0.32} & 0.81 & 0.54 & 0.91 & 0.95 \\
\midrule
\textbf{\methodname{} (Ours, Waymo)}
 & \textbf{0.632} & \textbf{0.27} & \underline{0.23} & \textbf{0.51} & \underline{0.46} & \underline{0.27} & \textbf{0.85} & \textbf{0.57} & \textbf{0.93} & \textbf{0.97} \\
\textbf{\methodname{} (Ours, nuPlan, zero-shot)}
 & \underline{0.625} & 0.22 & \underline{0.23} & \underline{0.51} & \underline{0.46} & 0.26 & \underline{0.84} & \underline{0.56} & \underline{0.93} & \underline{0.97} \\
\bottomrule
\end{tabular}%
}
\end{table}

The 2024 edition scores vehicles and vulnerable road users (VRUs) separately, since their dynamics and behavioral patterns differ, so we report a vehicle table and a VRU table following the SPACeR protocol. WOSAC builds a per-feature negative-log-likelihood of the logged outcome under the simulated rollout distribution and aggregates the kinematic, interactive, and map-based features under the official weights, normalized within each bucket, so realism reads as $0.20$ kinematic $+0.45$ interactive $+0.35$ map. For VRUs we drop the vehicle-only time-to-collision term and renormalize the remaining interactive weights within that bucket. The tables also report the minimum average displacement error (minADE) of the rollouts. The Demo-Free column records how much logged data each method requires: a check mark for methods that use no logged human data at all, an open circle for methods that use no demonstrations directly but depend on a separate reference policy trained on logged data, and a cross for methods trained directly on demonstrations. The \methodname{} entries come from one Waymo-trained checkpoint at epoch 12,000 and one nuPlan-trained checkpoint at epoch 4,000, each of which controls vehicles, pedestrians, and cyclists jointly. The SPACeR vehicle and VRU numbers come from two separately configured runs.

\begin{table}[t]
\centering
\caption{WOSAC 2024 vehicle sim agent evaluation on the filtered CAT-K validation subset. The realism, kinematic, interactive, and map columns are likelihood meta-scores (higher is better); minADE, collision, and off-road are rollout statistics (lower is better). The Demo-Free column uses \cmark{} for no logged human data, $\circ$ for an anchoring reference policy trained on logged data, and \xmark{} for direct demonstration training. Shaded rows are tokenized imitation-learning references; \textbf{bold} marks the best self-play entry per column and \underline{underline} the second best.}
\label{tab:wosac2024_veh}
\footnotesize
\setlength{\tabcolsep}{4pt}
\begin{tabular}{@{}lcccccccc@{}}
\toprule
Method
 & \makecell{Demo\\Free}
 & \makecell{Realism\\$\uparrow$}
 & \makecell{Kinematic\\$\uparrow$}
 & \makecell{Interactive\\$\uparrow$}
 & \makecell{Map\\$\uparrow$}
 & \makecell{minADE\\$\downarrow$}
 & \makecell{Coll.\\$\downarrow$}
 & \makecell{Off-rd.\\$\downarrow$} \\
\midrule
\rowcolor[gray]{0.93} SMART~\citep{wu2024smart}
 & \xmark & 0.720 & 0.450 & 0.725 & 0.870 & 1.84 & 0.170 & 0.130 \\
\rowcolor[gray]{0.93} CAT-K~\citep{zhang2025catk}
 & \xmark & 0.766 & 0.490 & 0.792 & 0.890 & 1.47 & 0.060 & 0.090 \\
\midrule
PPO~\citep{kazemkhani2025gpudrive}
 & \cmark & 0.710 & 0.327 & 0.751 & \underline{0.875} & 12.73 & 0.038 & 0.053 \\
HR-PPO~\citep{cornelisse2024hrppo}
 & $\circ$ & 0.716 & 0.341 & 0.756 & \textbf{0.880} & 12.25 & 0.044 & 0.053 \\
SPACeR~\citep{chang2026spacer}
 & $\circ$ & \textbf{0.741} & \underline{0.411} & 0.779 & \textbf{0.880} & \textbf{4.10} & 0.036 & 0.056 \\
\midrule
\textbf{\methodname{} (Ours, Waymo)}
 & \cmark & \underline{0.740} & \textbf{0.412} & \textbf{0.797} & 0.853 & \underline{6.07} & \textbf{0.006} & \textbf{0.037} \\
\textbf{\methodname{} (Ours, nuPlan, zero-shot)}
 & \cmark & 0.733 & 0.393 & \underline{0.795} & 0.847 & 7.92 & \underline{0.007} & \underline{0.039} \\
\bottomrule
\end{tabular}
\end{table}

\begin{table}[t]
\centering
\caption{WOSAC 2024 VRU sim agent evaluation on the filtered CAT-K validation subset. All columns are higher-is-better except minADE. The Demo-Free column uses \cmark{} for no logged human data, $\circ$ for an anchoring reference policy trained on logged data, and \xmark{} for direct demonstration training. \textbf{Bold} marks the best self-play entry per column and \underline{underline} the second best.}
\label{tab:wosac2024_vru}
\footnotesize
\setlength{\tabcolsep}{4pt}
\begin{tabular}{@{}lcccccc@{}}
\toprule
Method
 & \makecell{Demo\\Free}
 & \makecell{Realism\\$\uparrow$}
 & \makecell{Kinematic\\$\uparrow$}
 & \makecell{Interactive\\$\uparrow$}
 & \makecell{Map\\$\uparrow$}
 & \makecell{minADE\\$\downarrow$} \\
\midrule
PPO~\citep{kazemkhani2025gpudrive}
 & \cmark & 0.648 & 0.242 & 0.683 & 0.835 & 7.71 \\
HR-PPO~\citep{cornelisse2024hrppo}
 & $\circ$ & 0.668 & 0.285 & 0.700 & \underline{0.847} & 7.01 \\
SPACeR~\citep{chang2026spacer}
 & $\circ$ & \textbf{0.729} & \textbf{0.413} & \textbf{0.762} & \textbf{0.866} & \textbf{2.07} \\
\midrule
\textbf{\methodname{} (Ours, Waymo)}
 & \cmark & \underline{0.688} & 0.365 & \underline{0.738} & 0.808 & \underline{2.85} \\
\textbf{\methodname{} (Ours, nuPlan, zero-shot)}
 & \cmark & 0.686 & \underline{0.369} & 0.733 & 0.808 & 2.94 \\
\bottomrule
\end{tabular}
\end{table}

\paragraph{Discussion.}
Table~\ref{tab:wosac2023} reports WOSAC 2023 against Gigaflow and the demonstration-free baselines. Unlike Gigaflow, whose WOSAC entry drives pedestrians with a scripted IDM-like controller and applies its learned policy only to vehicles and cyclists, \methodname{} controls vehicles, pedestrians, and cyclists through one unified heterogeneous policy, which makes it suitable for unified, interactive traffic-scene simulation. \methodname{} edges Gigaflow on overall realism (0.632 versus 0.619) and leads on linear and angular speed, time-to-collision, road-edge distance, off-road, and collision, while trailing on distance-to-object and the acceleration kinematics.

Tables~\ref{tab:wosac2024_veh} and~\ref{tab:wosac2024_vru} report WOSAC 2024 against recent self-play methods. On vehicles, \methodname{} nearly matches SPACeR on the realism composite (0.740 versus 0.741), slightly exceeds its kinematic score (0.412 versus 0.411), and leads the self-play entries on interactive likelihood (0.797 versus 0.779). Its lower map likelihood (0.853 versus 0.880) accounts for the remaining composite gap. \methodname{} also posts the lowest collision and off-road rates of any method in the table, with a higher minADE than SPACeR. On VRUs, \methodname{} reaches a realism composite of 0.688 that surpasses PPO (0.648) and HR-PPO (0.668) and trails SPACeR (0.729), with a VRU minADE (2.85) close to SPACeR (2.07) and far below PPO and HR-PPO. \methodname{} reaches this without demonstrations, without an anchoring reference policy, and without logged-trajectory initialization, where SPACeR and HR-PPO each rely on a reference policy trained on logged data.

The same recipe transfers across datasets. A heterogeneous policy trained only on nuPlan map geometry, evaluated zero-shot on the Waymo WOSAC protocol with no Waymo training, nearly matches the Waymo-trained policy: vehicle realism $0.733$ versus $0.740$ and VRU realism $0.686$ versus $0.688$ under the 2024 protocol, and overall realism $0.625$ versus $0.632$ under the 2023 protocol (Tables~\ref{tab:wosac2023}, \ref{tab:wosac2024_veh}, and~\ref{tab:wosac2024_vru}). This near-parity echoes the transfer study of Section~\ref{subsec:generalization} (Figure~\ref{fig:transfer-matrix}), where realism tracks the domain-randomization scheme rather than the training source; the WOSAC comparison here measures that same cross-dataset transfer under the full leaderboard protocol rather than the restricted transfer-matrix protocol.

The two benchmarks together support our central claim: \methodname{} is not merely a fast simulator but a training substrate that yields policies competitive as both ego driving policies (Table~\ref{tab:val14}) and sim agents (Tables~\ref{tab:wosac2023}, \ref{tab:wosac2024_veh}, and~\ref{tab:wosac2024_vru}). A policy that closely mimics human kinematics may still fail on task metrics, while a policy can game val14 through overly cautious driving that diverges from human behavior; the two metrics therefore catch complementary failure modes.

\subsection{Generalization}
\label{subsec:generalization}

\begin{figure}[t]
\centering
\includegraphics[width=\linewidth]{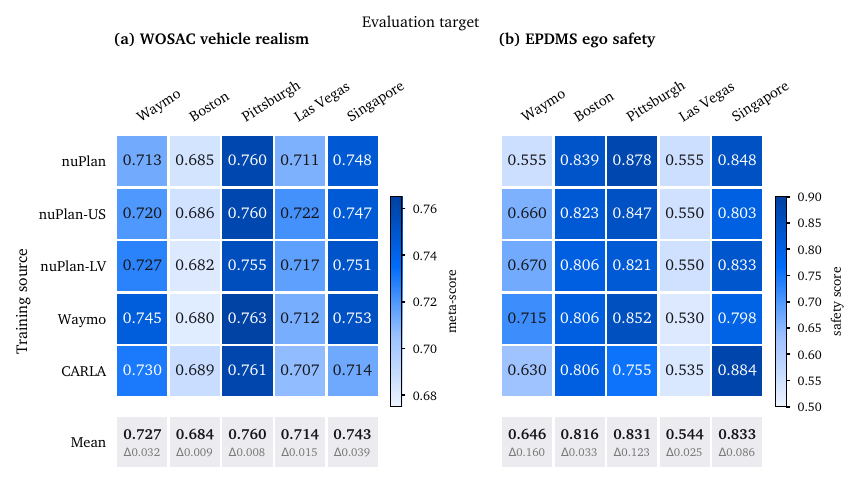}
\caption{\textbf{Transfer matrices across datasets and cities.} Each \methodname{} policy trains on one source (row) and is scored zero-shot on each target (column), under (a) the WOSAC 2024 vehicle-realism meta-score and (b) the NAVSIM v2 EPDMS core ego-safety score, where darker is higher. Both panels band by column rather than by row, so a policy's score is set by the evaluation target and not by the source it trained on. The bottom strip of each panel gives the per-target mean with the across-source spread ($\Delta$, max minus min) below it. Row labels abbreviate the nuPlan US-cities and Las Vegas training splits; CARLA is a source only, since its maps carry no logged trajectories to score.}
\label{fig:transfer-matrix}
\end{figure}

A central claim of our domain-randomization approach is that policy robustness derives from the randomization scheme rather than from memorizing dataset-specific patterns. We test this directly with a transfer study: a separate \methodname{} policy is trained on each data source, each for \textit{fewer training steps} than the main benchmark policies of Sections~\ref{subsec:val14} and~\ref{subsec:wosac}, and then evaluated, \emph{zero-shot} with no fine-tuning on the target, on every other dataset and nuPlan city. We read each policy through two complementary lenses, distributional \emph{realism} scored with the WOSAC vehicle protocol and ego \emph{safety} scored with the NAVSIM v2 Extended PDM Score (EPDMS)~\citep{cao2025navsimv2}. Both lenses share the same five checkpoints and five targets, so together they yield a pair of transfer matrices whose rows are the training source and whose columns are the evaluation target (Figure~\ref{fig:transfer-matrix}).

\paragraph{Setup.}
The realism matrix scores each cell under the WOSAC 2024 protocol restricted to vehicles, reporting the vehicle realism meta-score~\citep{montali2023wosac}. The metric compares the distribution of simulated vehicle features against the logged ground truth over an 80-frame future horizon under 32 stochastic closed-loop rollouts, weighting kinematic, interactive, and map-based feature families. Agents are initialized from the logged state but pursue \emph{randomly generated} goals, so the policy never observes the held-out logged destination. The safety matrix scores the same checkpoints with the NAVSIM v2 EPDMS~\citep{cao2025navsimv2, dauner2023pdm}, reporting the core safety score, the minimum of the at-fault-filtered no-collision, drivable-area, driving-direction, and time-to-collision sub-metrics, so a single safety failure caps the cell. In this setup the policy controls only the ego vehicle while every other agent replays its logged trajectory. This isolates ego transfer behavior but leaves the surrounding traffic \emph{non-reactive}, so the EPDMS score here measures ego safety against replayed agents rather than the fully reactive closed-loop driving of the nuPlan val14 benchmark in Table~\ref{tab:val14}. Both matrices restore each checkpoint's own training configuration and vary only the target maps, and each cell averages about 100 scenarios, which supports relative comparison rather than leaderboard-grade absolutes. Waymo realism cells use the official WOSAC subject set, while the nuPlan cities have no official subject set and so score all valid logged vehicles, giving WOSAC-2024-\emph{style} rather than official-subject numbers.

\paragraph{Cross-dataset transfer.}
The first question is whether a policy pays a penalty for being evaluated away from the dataset it trained on. It does not, under either lens. Reading down any column of Figure~\ref{fig:transfer-matrix}, the score barely moves with the training source: the per-source mean realism spans only about $0.011$ across the five policies (from $0.720$ for the CARLA-trained policy to $0.731$ for the Waymo-trained one), and the per-source mean safety spans about $0.018$ (from $0.722$ to $0.740$), with no policy holding a consistent home-dataset advantage in either. A Waymo-trained policy scores $0.745$ realism on Waymo, yet a nuPlan-trained policy reaches $0.713$ there, and on nuPlan Singapore the Waymo-trained policy ($0.753$) edges the nuPlan-trained one ($0.748$). The CARLA-trained policy, whose training maps are just five synthetic towns, nonetheless lands in this same narrow band as the Waymo- and nuPlan-trained policies drawn from hundreds of thousands of scenes ($0.720$ realism and $0.722$ safety, within $0.011$ and $0.018$ of the best), so transfer quality tracks the randomization scheme rather than the size of the training corpus. Both panels read as near-uniform color within each column: the dominant variation runs across columns, a property of the target, not across rows.

\paragraph{Cross-city transfer.}
Within nuPlan the same pattern holds at the level of individual cities. The policy trained only on Las Vegas and the policy trained on the US cities both transfer to held-out cities with scores indistinguishable from the policy trained on the full nuPlan mix (Figure~\ref{fig:transfer-matrix}, bottom strips), even though Las Vegas supplies several times more training scenes than Boston or Singapore. Where the columns differ is between cities, not between sources: under realism, Boston sits lowest ($0.684$ mean) and Pittsburgh highest ($0.760$); under safety, Las Vegas is the uniformly low column ($0.544$ mean, spread only $0.025$ across sources) while the nuPlan cities score high ($0.82$ to $0.83$). The dominant training city thus earns no home advantage, and the safety score binds hardest on that very city. These are gaps of the target rather than of any policy: the two lenses disagree about which targets are hard, and the low realism on Boston and Pittsburgh is largely an artifact of road-edge geometry in the converted maps that every policy inherits equally, not a transfer penalty; on those two cities the off-road metric flags even logged human vehicles as off-road at rates several times the level seen elsewhere.

\paragraph{Emergent left-hand-traffic driving.}
The transfer matrices contain a built-in test of whether the policy memorizes a driving side or follows road structure. Four of the five training sources (Waymo, CARLA, the nuPlan US cities, and Las Vegas) contain only right-hand-traffic maps, yet nuPlan Singapore, a left-hand-traffic country, is among the highest-scoring targets under both metrics (a $0.743$ realism column mean and a $0.833$ safety column mean). The policy trained on the US cities, which never sees a left-hand-traffic map, scores $0.747$ realism on Singapore, matching the $0.748$ of the policy trained on the full nuPlan mix, whose training pool is the only source that includes Singapore itself. A policy that had memorized right-hand travel would collapse here; instead its realism and safety are undiminished, which indicates that the learned behavior is to follow the local lane topology rather than a fixed directional convention. We attribute this to kinematic domain randomization combined with reward-conditioned training, which reward progress along the road structure the agent is placed on. As noted by \citet{wang2026nomad}, achieving such transfer without human demonstrations is a key indicator of genuine robustness.

\subsection{Qualitative Analysis}
\label{subsec:qualitative}

\begin{figure}[t]
\centering
\includegraphics[width=\textwidth]{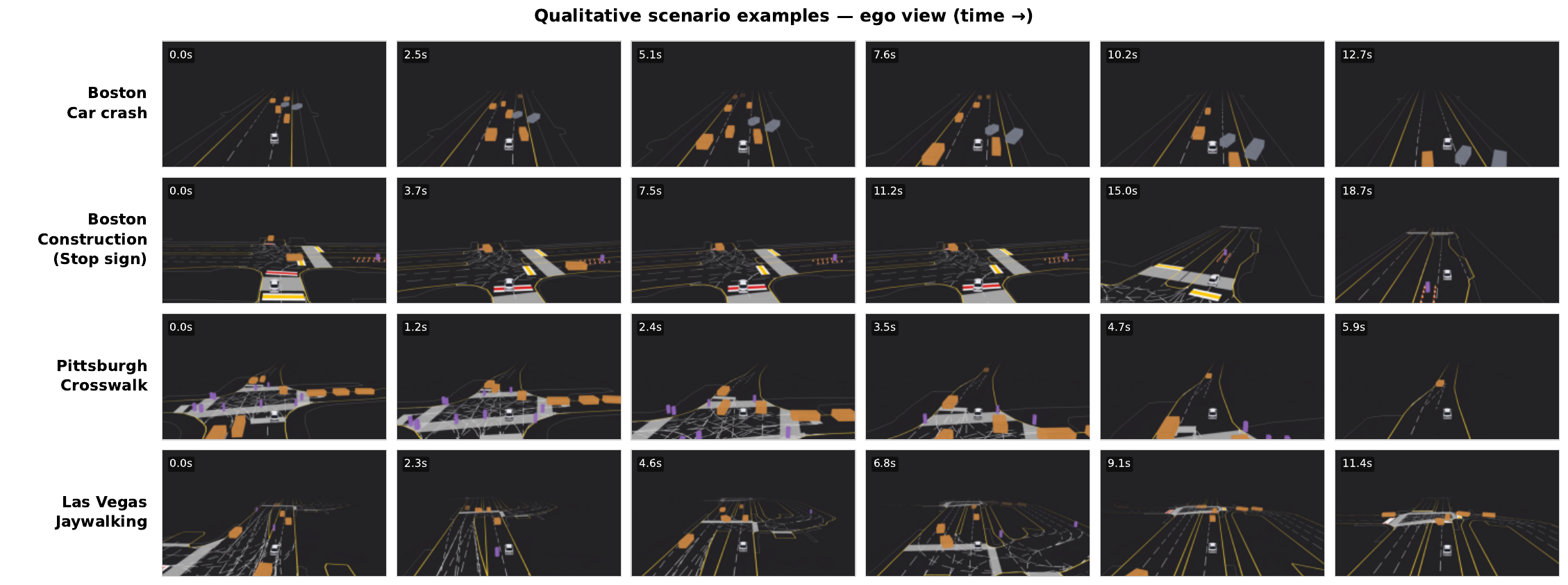}
\caption{\textbf{Qualitative ego-policy rollouts.} Four safety-critical scenario types and cities (top to bottom: a Boston car crash, a Boston stop-sign construction zone, a Pittsburgh crosswalk, and a Las Vegas jaywalking pedestrian). Each row is a single episode shown as a left-to-right time sequence in the ego view; the \methodname{} policy sustains progress while avoiding collisions and yielding to vulnerable road users.}
\label{fig:qualitative}
\end{figure}

Beyond aggregate metrics, we visualize policy behavior in the situations that most stress a driving policy. Figure~\ref{fig:qualitative} shows ego-view rollouts of a single \methodname{} policy across four safety-critical scenario types drawn from different cities: a car-crash encounter in Boston, a stop-sign construction zone in Boston, a pedestrian crosswalk in Pittsburgh, and a jaywalking pedestrian in Las Vegas. Each row reads left to right as the episode advances, with the policy-controlled ego agent navigating the hazard while reactive traffic and pedestrians evolve around it. The policy keeps progress through congestion, respects the stop-sign and crosswalk geometry, and yields to the jaywalking pedestrian rather than colliding. These behaviors are consistent with the quantitative gains in the No AF-Collision and TTC components of Table~\ref{tab:val14}. Video rollouts of these and additional scenarios are available on the \href{https://terra-applied.github.io/TerraZero}{project website}, covering procedurally generated long-tail scenarios for cars and trucks, heterogeneous multi-agent simulation, closed-loop driving on val14 and InterPlan, and traffic simulation on WOSAC.

%% file: sections/conclusion.tex
\section{Conclusion}
\label{sec:conclusion}

We have presented \methodname, a high-performance closed-loop driving simulator designed for self-play reinforcement learning at scale.
\methodname{} addresses a persistent tension in autonomous driving research: the simulators fast enough for RL-scale training have historically lacked the scenario fidelity needed for meaningful policy transfer, while feature-rich environments remain orders of magnitude too slow.
By combining a configurable C simulation engine, a procedural scenario generator, and a compute-efficient self-play recipe, \methodname{} achieves among the highest reported throughput for object-level, self-play-enabled driving simulators while supporting heterogeneous traffic (vehicles, pedestrians, cyclists), multiple dynamics models, traffic-rule enforcement, and multi-source map data from Waymo, nuPlan, and CARLA.

The system makes three concrete contributions: a fast, feature-rich object-level C simulator that retains full scenario fidelity at high throughput across consumer-grade, server-grade, and multi-GPU hardware; a procedural scenario generator that manufactures a vast space of scenarios from real-world maps rather than training on the logs directly; and a compute-efficient self-play recipe that trains every reported policy from scratch with zero human demonstrations, using no imitation and no logged trajectories, and generalizes zero-shot across cities and datasets. We measure driving performance on nuPlan val14 and InterPlan, and sim-agent realism on WOSAC.
The same recipe tops both the standard val14 benchmark and the interactive long-tail InterPlan benchmark, and outperforms other demonstration-free methods on WOSAC realism, so one stack yields both a realistic traffic simulator and a high-performance, robust planner.

\paragraph{Limitations.}
\methodname{} relies on high-definition maps with lane-level topology, traffic signal phase information, and intersection geometry; regions for which such maps are unavailable cannot be used as training or evaluation scenarios.
The object-level simulation abstraction does not model visual perception: policies operate on ground-truth structured features rather than camera or lidar inputs, so the system cannot directly train end-to-end perception-to-control pipelines.
While domain randomization over kinematic parameters improves robustness, the sim-to-real gap for physical vehicle dynamics (tire friction, suspension response, aerodynamic effects) remains an open challenge that randomization alone does not fully resolve.

\paragraph{Future Work.}
Several directions extend naturally from the current system.
\emph{City-scale simulation} would increase scenario duration and spatial extent to support long-horizon planning across entire urban networks, moving beyond the intersection-level episodes that currently dominate training.
\emph{Expanded agent types} (emergency vehicles, construction equipment, electric scooters, and other underrepresented road users) would enrich interaction dynamics and further stress-test policy generalization.
\emph{Sim-to-real transfer experiments} leveraging the kinematic domain randomization framework would provide direct evidence of whether policies trained in \methodname{} transfer to physical platforms, closing the loop between simulation and deployment.
Finally, \emph{integration with visual perception pipelines} (replacing ground-truth observations with learned representations from camera or lidar inputs) would bridge the gap between object-level and end-to-end autonomous driving, enabling \methodname{} to serve as a training environment for full-stack driving systems.

%% file: sections/appendix.tex

\section{Simulation Constants}
\label{app:constants}

Table~\ref{tab:sim-constants} lists the structural parameters that define the simulation environment's capacity and behavior.

\begin{table}[ht]
\centering
\caption{Simulation constants. Compile-time limits govern memory allocation; runtime parameters are configurable per experiment.}
\label{tab:sim-constants}
\small
\begin{tabularx}{\textwidth}{l >{\raggedright\arraybackslash}X c}
\toprule
\textbf{Category} & \textbf{Parameter} & \textbf{Value} \\
\midrule
\multicolumn{3}{l}{\textit{Observations}} \\
& Max observed partner agents & 20 \\
& Max observed road segments & 200 \\
& Max observed traffic entities & 16 \\
& Default partner observation radius & 50\,m \\
& Default traffic observation radius & 100\,m \\
\midrule
\multicolumn{3}{l}{\textit{Time}} \\
& Simulation timestep ($\Delta t$) & 0.1\,s \\
& Policy timestep & 0.1\,s \\
& Episode length & 256 steps (25.6\,s) \\
\midrule
\multicolumn{3}{l}{\textit{Map Structure}} \\
& Max phases per intersection & 8 \\
& Max signals per intersection & 16 \\
& Max stop lines per intersection & 16 \\
& Max travel directions per lane & 32 \\
& Z-separation threshold (multi-level filtering) & 2.5\,m \\
\bottomrule
\end{tabularx}
\end{table}

\section{Reward Function Details}
\label{app:rewards}

Throughout this appendix, a \emph{Used by} tag records whether a setting applies to both reported policies of Section~\ref{subsec:experimental-setup} (Both), only the planner (Planner), or only the heterogeneous sim agent (Sim agent).

\begin{table}[ht]
\centering
\caption{Reward terms and their per-type weights. Each term is computed per timestep per agent; entries are the per-agent randomization range $[\min, \max]$ or a fixed value, and a dash marks a term disabled for that type. The planner controls vehicles only and so uses the Vehicle column, while the heterogeneous sim agent uses all three. Goal progress and target speed are held at zero across both reported policies and are omitted, so the full parameter vector has 23 entries (Appendix~\ref{app:observations}). The auxiliary thresholds $\delta_{\text{goal}}$, $v_{\text{goal}}$, $b_{\text{center}}$, and $v_{\text{speed\_lim}}$ are defined below.}
\label{tab:reward_details}
\small
\begin{tabularx}{\textwidth}{l >{\raggedright\arraybackslash}X l ccc}
\toprule
Term & Formula & Category & Vehicle & Pedestrian & Cyclist \\
\midrule
Goal & $+\alpha_{\text{goal}}$ if $\|p - p_g\| < \delta_{\text{goal}}$ and $v < v_{\text{goal}}$ & Navigation & 1.0 & 1.0 & 1.0 \\
Collision & $-(\alpha_{\text{col}} + \gamma_{\text{col}} \cdot v)$ & Safety & $[0.0, 3.0]$ & $[0.5, 2.0]$ & $[1.0, 3.0]$ \\
Boundary & $-\alpha_{\text{bound}}$ if off drivable area & Safety & $[0.0, 3.0]$ & --- & $[0.5, 2.0]$ \\
Comfort & $-\alpha_{\text{comf}} \cdot n_{\text{viol}}$ & Comfort & $[0.0, 0.1]$ & --- & --- \\
Lane Align & $-\alpha_{\text{align}} \cdot |\theta_f|$ & Lane & $[2.5e{-}4, 2.5e{-}2]$ & --- & $[5e{-}3, 2e{-}2]$ \\
Lane Center & $-\alpha_{\text{center}} \cdot |x_f + b_{\text{center}}|$ & Lane & $[2.5e{-}4, 7.5e{-}3]$ & --- & --- \\
Velocity & $+\alpha_{\text{vel}} \cdot \Delta t \cdot v / v_{\max}$ & Speed & $2.5e{-}3$ & --- & --- \\
Vel Align & $-\alpha_{\text{va}} \cdot |\Delta v|$ & Speed & $[0.0, 1.0]$ & --- & $[0.25, 0.75]$ \\
Reverse & $-\alpha_{\text{rev}}$ if $v < 0$ & Speed & $[2.5e{-}4, 7.5e{-}3]$ & --- & --- \\
Timestep & $+\alpha_{\text{step}}$ (survival) & Speed & $2.5e{-}5$ & --- & --- \\
Stop Line & $-\alpha_{\text{stop}}$ if violating & Compliance & 5.0 & --- & 5.0 \\
Red Light & $-(\alpha_{\text{red}} + \gamma_{\text{col}} \cdot v)$ & Compliance & 3.0 & --- & 3.0 \\
Road Incursion & $-\alpha_{\text{incur}} \cdot \Delta t$ if off-road and off-crosswalk & Safety & --- & $[0.5, 2.0]$ & --- \\
Speed Limit & $-\alpha_{\text{slim}} \cdot \Delta t \cdot (v - v_{\text{lim}}) / v_{\max}$ if $v > v_{\text{lim}}$ & Speed & --- & $[0.5, 2.0]$ & $[0.25, 1.0]$ \\
Edge Preference & $+\alpha_{\text{edge}} \cdot \Delta t \cdot \mathrm{clamp}(x_f / w_{\text{lane}})$ & Lane & --- & --- & $[0.01, 0.05]$ \\
\bottomrule
\end{tabularx}
\end{table}

\paragraph{Variable definitions.}
\begin{itemize}[leftmargin=*, itemsep=1pt, topsep=2pt]
    \item $p, p_g$: agent position and goal position in world frame.
    \item $\delta_{\text{goal}}$: goal region radius (default 2.0\,m; randomizable in $[2, 12]$\,m).
    \item $v_{\text{goal}}$: maximum speed for goal acceptance (default 3.0\,m/s). An agent must be within $\delta_{\text{goal}}$ \emph{and} traveling below $v_{\text{goal}}$ to register goal achievement, preventing drive-through exploits.
    \item $\gamma_{\text{col}}$: collision speed scale (default 0.1), controlling how strongly speed amplifies the collision and red-light penalties. At $v = 20$\,m/s the penalty roughly doubles compared to a stationary collision.
    \item $n_{\text{viol}}$: count of comfort threshold violations per timestep. Four thresholds are checked independently: $|a_{\text{long}}| > 3.0$\,m/s$^2$, $|a_{\text{lat}}| > 3.0$\,m/s$^2$, $|j_{\text{long}}| > 5.0$\,m/s$^3$, $|j_{\text{lat}}| > 5.0$\,m/s$^3$. The penalty is proportional to the number of violations ($n_{\text{viol}} \in \{0,1,2,3,4\}$), not their magnitude. The thresholds are per agent type; the values above are for vehicles, while pedestrians use $(1.5, 1.0, 12.0, 5.0)$ and cyclists $(3.0, 1.5, 40.0, 18.0)$ for $(|a_{\text{long}}|, |a_{\text{lat}}|, |j_{\text{long}}|, |j_{\text{lat}}|)$.
    \item $\theta_f$: heading deviation from the nearest lane's travel direction (in radians).
    \item $x_f$: signed lateral offset from lane center (in meters). $b_{\text{center}}$ is a configurable center bias (default 0.0) that shifts the preferred lateral position, e.g., for right-of-lane preference.
    \item $v_{\max}$: maximum velocity after kinematic scaling, $v_{\max} = 20.0 \cdot c_{\text{vel}}$\,m/s. The base clip is 20.0\,m/s; the per-agent coefficient $c_{\text{vel}} \in [0.5, 1.5]$ modulates it.
    \item $v_{\text{lim}}$: speed-limit cap for the sim agent's speed-limit term, resolved per lane from the posted limit when one is available.
    \item $v_{\text{speed\_lim}}$: per-type speed-limit parameter sampled per agent (pedestrians $[1.5, 2.5]$, cyclists $[5.0, 8.0]$; disabled for vehicles).
    \item $\Delta t$: simulation timestep (0.1\,s). The velocity and speed-based rewards are multiplied by $\Delta t$ to produce per-step rather than per-second rewards.
\end{itemize}

\paragraph{Reward parameter randomization.}
Each of the 23 reward parameters supports three modes: a \textbf{fixed scalar} (constant across all agents and episodes), a \textbf{$[\min, \max]$ pair} (uniformly sampled per agent at episode start), or \textbf{null} (excluded from reward computation, and from the observation vector when null for every agent type). The per-type weights are those of Table~\ref{tab:reward_details}: the planner uses the Vehicle column, while the heterogeneous sim agent uses all three.

Sampled parameters are normalized to $[-1, 1]$ via $\hat{\alpha}_i = 2 (\alpha_i - \alpha_i^{\min}) / (\alpha_i^{\max} - \alpha_i^{\min}) - 1$ before inclusion in the ego observation vector. Fixed parameters (where $\alpha_i^{\min} = \alpha_i^{\max}$) normalize to 0.

\section{Observation Space Details}
\label{app:observations}

The observation space is organized into four groups, each encoded via Deep Sets~\citep{zaheer2017deepsets} (a shared MLP per entity, followed by max-pooling to produce a fixed-length vector regardless of entity count). All observations are computed in the \textbf{ego agent's local coordinate frame} and zero-padded when fewer than the maximum number of entities are present. Each feature can be independently enabled or disabled, and the network input resizes to the enabled set.

\paragraph{Ego-centric coordinate transform.}
All spatial features (positions, headings, velocities) are expressed relative to the ego agent. Given a world-frame displacement $(\Delta x, \Delta y)$ and the ego agent's heading unit vector $(\cos\psi, \sin\psi)$, the ego-frame coordinates are:
\begin{equation}
\label{eq:ego_transform}
x_{\text{ego}} = \Delta x \cos\psi + \Delta y \sin\psi, \qquad
y_{\text{ego}} = -\Delta x \sin\psi + \Delta y \cos\psi,
\end{equation}
where $x_{\text{ego}}$ points forward along the ego heading and $y_{\text{ego}}$ points left. This transform is applied to goal positions, partner positions, road segment midpoints, and traffic entity positions.

The active reward weight parameters are included in the ego observation vector (\textbf{reward-conditioned observations}), enabling a single policy to generalize across reward configurations: adjusting the weight vector at inference time changes driving behavior without retraining. The planner exposes 19 of the 23 parameters (its four vulnerable-road-user terms are null), while the heterogeneous sim agent exposes all 23, since no parameter is null across its three agent types.

\begin{table}[ht]
\centering
\caption{Ego features for the planner (1 agent). Features marked with $\dagger$ are optional and controlled by configuration flags. The heterogeneous sim agent exposes all 23 reward parameters in its \texttt{alpha\_params} block (none are null across its three types).}
\label{tab:obs_ego}
\small
\begin{tabular}{lcl}
\toprule
Feature & Dims & Computation \\
\midrule
\texttt{agent\_type} & 1 & Categorical index: vehicle=1, pedestrian=2, cyclist=3 \\
\texttt{rel\_goal\_pos} & 2 & Ego-frame goal position (Eq.~\ref{eq:ego_transform}), scaled $\times 0.005$ \\
\texttt{goal\_dropout\_flag} & 1 & Binary: 1 when the goal is hidden this step (goal dropout) \\
\texttt{state\_dropout\_flag} & 1 & Binary: 1 when ego state features are dropped (robustness) \\
\texttt{speed} & 1 & $\operatorname{sign}(\mathbf{v}\cdot\hat{\mathbf{h}}) \sqrt{v_x^2 + v_y^2} \,/\, 100$ (signed; negative when reversing) \\
\texttt{vehicle\_dims} & 2 & (width / 15, length / 30) \\
\texttt{collision\_state} & 1 & Binary: 1 if currently in collision \\
\texttt{steering\_angle} & 1 & $\delta_{\text{steer}} / \pi$ \\
\texttt{acceleration} & 2 & $(a_{\text{long}} / 5, \; a_{\text{lat}} / 4)$ \\
\texttt{stop\_sign\_state}$^\dagger$ & 3 & One-hot: (not in region, must stop, cleared); on with traffic rules \\
\texttt{alpha\_params}$^\dagger$ & 19 & Active reward params normalized to $[-1, 1]$ (Sec.~\ref{app:rewards}) \\
\texttt{kinematic\_params}$^\dagger$ & 4 & $(c_{\text{throttle}}, c_{\text{steer}}, c_{\text{acc}}, c_{\text{vel}})$ normalized \\
\bottomrule
\end{tabular}
\end{table}

The engine also implements ego features that are disabled in the default configuration: lane state $(\theta_f/\pi,\, x_f/2.5)$, topology distance to goal, local road curvature, lateral velocity, and yaw rate.

\begin{table}[ht]
\centering
\caption{Partner features (nearest agents within the observation radius).}
\label{tab:obs_partner}
\small
\begin{tabular}{lcl}
\toprule
Feature & Dims & Computation \\
\midrule
\texttt{rel\_pos} & 2 & Ego-frame position (Eq.~\ref{eq:ego_transform}), scaled $\times 0.02$ \\
\texttt{vehicle\_dims} & 2 & (width / 15, length / 30) \\
\texttt{rel\_heading} & 2 & $(\cos\Delta\psi, \sin\Delta\psi)$ relative to ego heading \\
\texttt{speed} & 1 & $\sqrt{v_x^2 + v_y^2} \,/\, 100$ \\
\texttt{actor\_type} & 1 & Categorical: vehicle / pedestrian / cyclist \\
\texttt{acceleration}$^\dagger$ & 2 & $(a_{\text{long}} / 5, \; a_{\text{lat}} / 4)$; disabled by default \\
\midrule
\textbf{Total per partner} & \textbf{8} & \\
\bottomrule
\end{tabular}
\end{table}

Partners are the closest agents within the observation radius, filtered by elevation so that agents on a different road level are ignored.

\begin{table}[ht]
\centering
\caption{Road features (nearest road segments).}
\label{tab:obs_road}
\small
\begin{tabular}{lcl}
\toprule
Feature & Dims & Computation \\
\midrule
\texttt{rel\_pos} & 2 & Ego-frame midpoint of segment, scaled $\times 0.02$ \\
\texttt{segment\_dims} & 2 & (length / 100, width / 100) \\
\texttt{rel\_angle} & 2 & $(\cos\Delta\phi, \sin\Delta\phi)$: segment direction in ego frame \\
\texttt{road\_type} & 1 & Categorical: lane=0, line=1, edge=2 \\
\midrule
\textbf{Total per segment} & \textbf{7} & \\
\bottomrule
\end{tabular}
\end{table}

Road segments fill a fixed budget that prioritizes lane centerlines and road edges, the geometry most relevant to routing and to the off-road boundary, over lane-boundary markings, so dense markings cannot crowd the essential geometry out of the budget. Segments are filtered by elevation to handle multi-level road structures (bridges, overpasses).

\begin{table}[ht]
\centering
\caption{Traffic features (nearby traffic entities).}
\label{tab:obs_traffic}
\small
\begin{tabular}{lcl}
\toprule
Feature & Dims & Computation \\
\midrule
\texttt{entity\_type} & 1 & 0 = stop line, 1 = traffic light \\
\texttt{rel\_pos} & 3 & Ego-frame 3D position, scaled $(0.02, 0.02, 0.1)$ \\
\texttt{signal\_state} & 4 & One-hot: (red, yellow, green, off/unknown) \\
\texttt{stop\_line\_endpoints} & 4 & Ego-frame stop-line segment endpoints, scaled $\times 0.02$ \\
\midrule
\textbf{Total per entity} & \textbf{12} & \\
\bottomrule
\end{tabular}
\end{table}

Traffic entities are the stop lines nearest the ego, selected by ascending distance up to the budget, so the ego's own approaching stop line, its sole channel to the signal state, is never crowded out on dense maps.

\section{Agent Initialization}
\label{app:agent-init}

At each episode reset, the simulation initializes agents through a multi-step procedure that composes scenario data with several randomization mechanisms to produce diverse, challenging traffic configurations. Algorithm~\ref{alg:agent-init} summarizes the full initialization sequence.

\begin{algorithm}[ht]
\caption{Agent initialization at episode reset.}
\label{alg:agent-init}
\begin{algorithmic}[1]
\STATE Load terrabin scenario (map geometry, logged trajectories, signal phases)
\STATE Compute world-frame mean $({\bar x}, {\bar y})$ and center all coordinates
\STATE Initialize grid map and lane connectivity graph
\IF{random initialization}
    \STATE \textbf{Random placement}: sample agent count, place on lanes (Sec.~\ref{app:random-placement})
\ELSE
    \STATE \textbf{Log initialization}: load positions from recorded trajectories
\ENDIF
\STATE \textbf{Role assignment}: assign policy-controlled, log-replay, or IDM roles (Sec.~\ref{app:role-assignment})
\FOR{each active agent}
    \STATE Sample kinematic parameters $(c_{\text{throttle}}, c_{\text{steer}}, c_{\text{acc}}, c_{\text{vel}})$ (Sec.~\ref{app:kin-random})
    \STATE Sample reward parameters $(\alpha_1, \ldots, \alpha_{23})$ (Sec.~\ref{app:rewards}, Table~\ref{tab:reward_details})
    \STATE Initialize velocity and heading from trajectory data or defaults
    \STATE Reset collision state, metrics, and displacement tracking
\ENDFOR
\STATE \textbf{Goal assignment}: set navigation goals from scenario data or topology sampling (Sec.~\ref{app:goal-assignment})
\end{algorithmic}
\end{algorithm}

\subsection{Random Agent Placement}
\label{app:random-placement}

When random initialization is enabled, the agent count is sampled from a configurable range whose upper bound may exceed the number of agents in the source scenario's logged data, which lets scenes pack denser than any naturalistic recording. Agents are then placed onto the lane network under rejection sampling: each placement starts on a lane, aligned with its travel direction and at low speed, and is retried until it is collision-free and on the road. Lanes inside signalized intersections are excluded so agents do not begin mid-intersection. Bounding-box dimensions are sampled independently per agent from type-specific uniform ranges.

\subsection{Behavioral Role Assignment}
\label{app:role-assignment}

After placement, each agent is assigned a behavioral role based on the configured control mode, which selects the agent types under policy control: from the ego vehicle alone, through vehicles only (the planner setting), up to every vehicle, pedestrian, and cyclist (the sim-agent setting). Candidates are filtered so that only agents with a valid lane and a sufficiently distant goal enter policy control, ensuring a non-trivial navigation task; the remaining agents follow recorded trajectories or rule-based controllers as background traffic.

\subsection{Hybrid Initialization}
\label{app:hybrid-init}

Hybrid initialization provides a middle ground between fully random and fully log-based initialization. A configurable probability $p_{\text{random}} \in [0, 1]$ controls the per-environment decision: with probability $p_{\text{random}}$, agents in that environment are randomly placed; otherwise, they are initialized from logged trajectories at timestep~0. When initialized from logs, agents begin at their recorded position, heading, and velocity but immediately transition to policy control: the log provides only the initial condition, not the ongoing behavior.

This mode is particularly useful for bootstrapping training: logged initial conditions place agents in coherent, mid-traffic states (e.g., at highway speed in a merge lane) that random placement would rarely produce, accelerating the policy's exposure to interactive situations, at the cost of the state diversity that random placement provides.

\subsection{Kinematic Parameter Randomization}
\label{app:kin-random}

At episode start, four multiplicative coefficients are sampled per agent and held constant for the episode's duration. Each coefficient scales the corresponding dynamics parameter:

\begin{table}[ht]
\centering
\caption{Kinematic randomization coefficients, each randomized per agent between $0.5$ and $1.5$.}
\label{tab:kin-params}
\small
\begin{tabular}{lll}
\toprule
\textbf{Coefficient} & \textbf{Scales} & \textbf{Effect across the range} \\
\midrule
$c_{\text{throttle}}$ & Longitudinal jerk input & More/less responsive acceleration \\
$c_{\text{steer}}$ & Lateral jerk/steering input & Tighter/looser turning \\
$c_{\text{acc}}$ & Max longitudinal acceleration clip & Higher/lower accel.\ limits \\
$c_{\text{vel}}$ & Max velocity clip ($v_{\max} = 20 \cdot c_{\text{vel}}$\,m/s) & Faster/slower max speed \\
\bottomrule
\end{tabular}
\end{table}

Velocity and acceleration bounds, scaled by the respective coefficients, are applied after the dynamics integration step, ensuring physical plausibility regardless of the sampled coefficients.

All four coefficients are included in the ego observation vector (Table~\ref{tab:obs_ego}), normalized to $[-1, 1]$ using this range, so the policy can condition its behavior on the current dynamics regime online.

\subsection{Goal Assignment}
\label{app:goal-assignment}

Each agent requires a navigation goal for the goal-achievement reward and the goal-relative observation. Goals are assigned differently depending on the initialization mode:

\paragraph{Log-based goals.}
When using logged trajectories, the goal is set to the agent's final recorded position in the source scenario, verified to be reachable via the lane connectivity graph before the scenario enters the training pool.

\paragraph{Random goal generation.}
When random initialization is used, or when an agent that reached its goal receives a new one, goals are sampled by a forward walk along the lane connectivity graph, following random candidate lanes at junctions, up to a target arc length drawn from a bounded range. An alignment filter rejects candidate goals that fall behind the agent, and an agent for which no aligned goal can be found is removed from the scene to avoid training on degenerate configurations. The sampling bounds and filter thresholds are configurable, and we omit their values.

\section{Training Configuration}
\label{app:hyperparameters}

Table~\ref{tab:hyperparams} lists the training hyperparameters shared by all experiments unless otherwise noted, and Table~\ref{tab:policy-configs} summarizes where the two reported policies, the planner and the heterogeneous sim agent, differ. Simulation-level constants shared by both policies (episode length, decision clock, and observation budgets) are those of Table~\ref{tab:sim-constants}.

\begin{table}[ht]
\centering
\caption{Training hyperparameters. Values are shared by both reported policies; the settings on which the planner and the sim agent differ are collected in Table~\ref{tab:policy-configs}.}
\label{tab:hyperparams}
\small
\begin{tabular}{llc}
\toprule
Category & Parameter & Value \\
\midrule
\multicolumn{3}{l}{\textit{PPO}} \\
& Learning rate & $5 \times 10^{-4}$ \\
& Learning rate annealing & Linear decay \\
& Discount factor ($\gamma$) & 0.99 \\
& GAE $\lambda$ & 0.95 \\
& PPO clip coefficient ($\epsilon$) & 0.2 \\
& Value function clip coefficient & None (unclipped) \\
& Entropy coefficient & 0.01 \\
& Value function coefficient & 0.5 \\
& Max gradient norm & 0.5 \\
& Update epochs per rollout & 2 \\
\midrule
\multicolumn{3}{l}{\textit{V-trace}} \\
& V-trace enabled & True \\
& $\bar{\rho}$ (IS ratio clip) & 1.0 \\
& $\bar{c}$ (trace coefficient clip) & 1.0 \\
\midrule
\multicolumn{3}{l}{\textit{PopArt}} \\
& EMA decay (per step) & 0.9997 \\
& Minimum $\sigma$ & $10^{-4}$ \\
\midrule
\multicolumn{3}{l}{\textit{Priority Sampling}} \\
& Priority exponent ($\alpha$) & 0.85 \\
& IS correction $\beta_0$ & 0.85 \\
& $\beta$ annealing & Linear, to $\approx 0.978$ \\
\midrule
\multicolumn{3}{l}{\textit{Optimizer}} \\
& Optimizer & Adam \\
& Adam $\beta_1$ & 0.9 \\
& Adam $\beta_2$ & 0.999 \\
& Adam $\epsilon$ & $10^{-8}$ \\
\midrule
\multicolumn{3}{l}{\textit{Architecture}} \\
& Num minibatches & 16 \\
& Total agents per GPU & 512 \\
& Environments per GPU & 32 \\
& Vectorized batch size & 4 \\
\bottomrule
\end{tabular}
\end{table}

\begin{table}[ht]
\centering
\caption{Configuration of the two reported policies, grouped by topic. Shared optimization hyperparameters are those of Table~\ref{tab:hyperparams}; the reward and kinematic randomization ranges are detailed in Appendices~\ref{app:rewards} and~\ref{app:kin-random}, and the rule-based road users in Section~\ref{subsec:npcs}.}
\label{tab:policy-configs}
\small
\begin{tabularx}{\textwidth}{l >{\raggedright\arraybackslash}X >{\raggedright\arraybackslash}X}
\toprule
& \textbf{Planner} & \textbf{Sim agent} \\
\midrule
\multicolumn{3}{l}{\textit{Policy}} \\
Architecture & Compact MLP, ${\sim}3.5$M parameters & Heterogeneous MLP with per-type action heads, ${\sim}6.7$M parameters \\
Controlled classes & Vehicles & Vehicles, pedestrians, cyclists \\
Action space & Discrete jerk grid & Per-type heads: jerk, unicycle, and compact bicycle \\
\midrule
\multicolumn{3}{l}{\textit{Scenario}} \\
Initialization & Random on-lane placement & Random on-lane placement for all classes \\
Goal dropout & $0.3$ & $0.3$ \\
Traffic signals & Christmas controller & Christmas controller; pedestrians exempt from signal rules \\
Road users & Mixed vehicle, pedestrian, and obstacle generators & Vehicle generators only; pedestrians and cyclists are policy-controlled \\
\midrule
\multicolumn{3}{l}{\textit{Domain randomization}} \\
Reward weights & Vehicle column of Table~\ref{tab:reward_details} & All three type columns of Table~\ref{tab:reward_details} \\
Kinematic coefficients & $[0.5, 1.5]$ on all four & Vehicles $[0.5, 1.5]$; cyclists narrower ranges; pedestrians fixed \\
\midrule
\multicolumn{3}{l}{\textit{Compute and data}} \\
Hardware & $16$ NVIDIA A100 80GB GPUs & $32$ NVIDIA A100 80GB GPUs \\
Map source & nuPlan & Waymo Open Motion Dataset \\
\bottomrule
\end{tabularx}
\end{table}